\documentclass[12pt]{article}
\usepackage{amsmath,amssymb,amsfonts,amsthm}
\usepackage{enumitem}
\usepackage[numbers,sort&compress]{natbib}
\usepackage{xurl}
\usepackage{caption}
\usepackage{booktabs}
\usepackage{tabularx}
\usepackage{longtable}
\usepackage{setspace}
\usepackage{graphicx}
\usepackage[letterpaper,margin=1in]{geometry}
\usepackage[colorlinks=true,linkcolor=blue,citecolor=blue,urlcolor=blue]{hyperref}

\setlist{nosep}
\providecommand{\tightlist}{\setlength{\itemsep}{0pt}\setlength{\parskip}{0pt}}

\newtheoremstyle{mlnote}%
  {\topsep}{\topsep}%
  {\itshape}{}%
  {\bfseries}{}%
  {\newline}%
  {\thmname{#1}\thmnumber{ #2}\ \thmnote{#3}}
\theoremstyle{mlnote}
\newtheorem{proposition}{Proposition}

\newtheorem{theorem}{Theorem}
\newtheorem{lemma}{Lemma}
\newtheorem{corollary}{Corollary}
\newtheorem{remark}{Remark}

\newcommand{\papertitle}{Certifiable inversion of resampling-induced spectral distortion in high-dimensional particle filters}
\newcommand{\keywords}[1]{\par\medskip\noindent\textbf{Keywords:} #1\par\medskip}
\newcommand{\funding}[1]{\section*{Funding}#1}
\newcommand{\data}[1]{\section*{Data availability}#1}
\newcommand{\suppdata}[1]{\section*{Supplementary material}#1}

\begin{document}
\title{\papertitle}
\author{\parbox{0.94\textwidth}{\centering
\large Deming Song$^1$, Chi-Kin Lam$^{1,*}$, Kefeng Li$^1$, Hui Yu$^2$, Xilin Li$^3$ and Huagang Lin$^3$\\[8pt]
\small $^1$Macao Polytechnic University, Rua de Lu\'is Gonzaga Gomes, Macao, China\\[3pt]
\small $^2$College of Textile Science and Engineering, Wuyi University, Jiangmen, China\\[3pt]
\small $^3$Guangdong Provincial Key Laboratory of Novel Polyamide 6 Functional Fiber Materials Research and Application, Jiangmen, China\\[5pt]
\small $^*$Corresponding author: \href{mailto:cklamsta@mpu.edu.mo}{cklamsta@mpu.edu.mo}}}
\date{}
\maketitle

\begin{abstract}
This paper formulates and solves the inverse problem of recovering the pre-resampling spectrum of a particle filter's ensemble covariance from the importance weights and the observed spectrum alone. Resampling controls weight degeneracy, but in high dimensions it also distorts that spectrum: the distortion is a free multiplicative convolution with a law computable from the weights, so the repair
is a deconvolution. We first extend the forward spectral law to random importance weights by conditioning on the weights together with the realized offspring counts rather than on the weights alone, which makes the distortion law deterministic at the observed data and gives the zero-atom mass of the limiting law a direct reading as an effective-dimension proxy for the second-order information the filter has lost. Second, a regularized free deconvolution inverts the distortion with an explicit error budget at a fixed resolution scale, and a computable forward residual certifies when the model behind it is violated. Injectivity makes the inverse identifiable but not stable, so the budget rests on a resolution-restricted condition number whose positivity we prove, and is vacuous below that resolution rather than concealed in a hypothesis. Third, the repaired spectrum enters only as the preconditioner of a Metropolis-adjusted rejuvenation move, so spectral error changes the mixing constants and not the invariant target, and the full cycle obeys a geometrically summable error recursion whose abstention term is governed by when abstentions fall rather than how many occur. Pre-registered simulations confirm the forward law, an order-of-magnitude reduction of spectral error, and the abstention law. The repair recovers eigenvalue laws, not eigenvectors.
\end{abstract}

\section{Introduction}\label{sec:1}

A particle filter can report a healthy effective sample size (ESS) while its covariance estimate has already collapsed. The ESS is a scalar summary of the weight distribution; the object that the filter's own propagation, its proposal adaptation and any downstream Gaussian summary actually consume is the spectrum of the ensemble second-moment matrix. In the proportional regime \(d/N\to c\) these two objects decouple. In the companion paper \cite{ref01} we showed what a single resampling step does to the spectrum: the limiting spectral distribution factorizes into a population channel and a count channel,
\[
\mu_S=H\boxtimes\mathcal{T}_c(\nu),
\]
with exact zero-atom masses, a rank-and-spectral-gap transition and exact second-moment costs. The present paper solves the associated inverse problem, which is ill-posed in a way that is specific to this setting and which we make precise.

The zero-atom mass \(m_0\) of the limiting law has an interpretation that we develop in Section~\ref{sec:5} and which motivates everything that follows. Asymptotically a fraction \(m_0\) of the limiting eigenvalue mass sits at zero, so only a fraction \(1-m_0\) of the spectrum carries nonzero second-order information, and \((1-m_0)d\) is an effective-dimension proxy for spectral preconditioning and condition-number control. Weight-only summaries such as the ESS do not determine the limiting spectral law or its zero-atom mass, so they cannot say how much was lost. This is a structural information gap, not a claim that the ESS never responds to degeneration: the monitor experiment of Section~\ref{sec:10.5} shows synchronized dynamics, and even when the ESS alarms it cannot identify the distortion law or say how to repair it.

The forward map from population spectra to observed spectra is injective, so two population spectra can produce two different observed spectra under one and the same weight vector. A diagnostic that is a function of the weights alone therefore takes the same value on both, which is the precise sense in which spectral information is necessary rather than convenient (Lemma~\ref{lem:weight-only}).

The inverse problem is harder than invertibility alone suggests. The forward map, a free multiplicative convolution with the distortion law, has no uniformly Lipschitz inverse on any class containing nearby atoms, so no error budget can be carried by the inverse itself. We therefore work at a resolution scale \(\theta\) and control a resolution-restricted condition number, whose positivity we prove; the budget is informative above \(\theta\) and vacuous below it, and we say so rather than
concealing the floor in a hypothesis.

The motivation for a repair carrying its own diagnostics comes from the contrast with linear filtering.
The Kalman filter \cite{ref36} propagates an error covariance, a computable certificate of its
uncertainty that drives the gain and bounds its precision. Particle filters have no counterpart. Our
contribution is to give them one, and to delimit when it is trustworthy. Raising \(N\) does not by
itself remove the distortion: in the proportional limit it changes \(c\) and hence the distortion law,
and the pre-registered limit \(W_1(H_0\boxtimes T,H_0)\) remains bounded away from zero over the range studied. A different scheme changes \(T\), which once known can still be divided out at a fixed resolution scale. The repair buys spectral accuracy at fixed \(N\), the budget that binds in high
dimensions.

The main theorems hold in an exogenous-weight generalized sample-covariance model (G1--G4), covering
exogenous importance weights, sample-split and cross-fitted weighting, and conditionally exogenous
local approximations (Section~\ref{sec:3.4}). Its boundary has three regions, located numerically in
Section~\ref{sec:10.3}: where the repair is certified, where the mismatch diagnostic stays dormant
because the distortion law is unchanged even though the weights are not exogenous, and where neither
applies. The fluctuation matrix is Gaussian in the main theorems and universality is left open. No
claim about end-to-end filtering accuracy is made beyond the stated spectral and invariance results.

\subsection{Prior work and what is missing}\label{sec:2}

\emph{Resampling and weight degeneracy.} Weight degeneracy and the effective sample size go back
to \cite{ref02}, with SMC methodology surveyed in \cite{ref03,ref33}, convergence theory in
\cite{ref05} and the Feynman--Kac framework in \cite{ref04}; resampling schemes are compared in
\cite{ref38,ref39}, and in high dimensions the particle filter collapses unless the ensemble grows
exponentially with an effective dimension \cite{ref40,ref41}. These analyses quantify resampling through
Monte Carlo variance and weight degeneracy; its spectral effect on high-dimensional ensembles is the
subject of \cite{ref01}, whose forward law we invert.

\emph{Rejuvenation moves and ensemble filters.} Resample--move \cite{ref06} applies a target-invariant
Markov kernel after resampling, formalized for SMC samplers in \cite{ref07} and embedded in particle
MCMC in \cite{ref08}; geometric ergodicity under drift and minorization is established in
\cite{ref30}, and adapting the proposal covariance to the empirical covariance is adaptive Metropolis
\cite{ref46}. Ensemble transform methods \cite{ref19}, optimal-transport formulations of the ensemble
Kalman filter \cite{ref20}, hybrid ensemble Kalman and particle filters \cite{ref21}, differentiable
resampling \cite{ref22}, the feedback particle filter \cite{ref23} and interacting-particle samplers
\cite{ref24,ref25} avoid the multiplicative count channel altogether. Our addition is a proposal
preconditioner whose spectrum has been corrected for resampling distortion, confining spectral error to
the mixing constants, and we keep standard resampling rather than replacing it.

\emph{Inverse problems and regularization.} The unknown here is a measure rather than a
finite-dimensional parameter, the forward map is a free multiplicative convolution with a law that is
itself estimated, and the inverse is injective but not Lipschitz. This places the problem within the
classical theory of ill-posed problems and regularization \cite{ref44,ref49,ref50}; the statistical and
Bayesian viewpoints \cite{ref51,ref52} apply as well, since the distortion law is estimated from the
weights and the repair carries both an error budget and a mismatch diagnostic. What replaces a
singular-value decomposition or a discrepancy principle is a resolution-restricted condition number, a
budget valid above that resolution, and a forward residual certifying model mismatch.

\emph{Spectrum estimation and free deconvolution.} The Marchenko--Pastur law \cite{ref28} and its
generalization \cite{ref29} describe the sample spectrum as a deformation of the population spectrum;
El Karoui \cite{ref14} inverts this relation, Ledoit and Wolf \cite{ref15} derive nonlinear shrinkage,
and \cite{ref16} reviews eigenvalue cleaning. Free deconvolution with a known factor is developed for
signal processing in \cite{ref12,ref13}, with consistent covariance-eigenvalue estimation in
\cite{ref42} and a survey in \cite{ref43}. Our inversion differs in that the factor is a random
count-law channel built from the weights and in that it carries an error budget and a mismatch
diagnostic; nonlinear shrinkage is the external baseline of Section~\ref{sec:10.2}.

The paper makes three contributions.

\begin{enumerate}[leftmargin=*]

\item

  \textbf{The conditioning object and what it costs.} That spectral law extends to random importance
  weights when one conditions on the weights together with the realized counts, and we delimit when
  this conditioning is legitimate. Conditioning on the weights alone leaves the count diagonal random.
  Because the forward map is injective, this also fixes what no weight-only diagnostic can see: the
  zero-atom mass of the resulting law quantifies the second-order dimension the filter has lost.

\item

  \textbf{Regularized inversion with an explicit budget.} A regularized free-deconvolution estimator is
  consistent with a local error budget at a fixed resolution scale. Injectivity gives no Lipschitz
  inverse, so the budget rests on a resolution-restricted condition number whose positivity we
  establish, together with a computable forward residual that certifies violation of the model.

\item

  \textbf{Diagnosed, target-preserving use.} The repaired spectrum is used only as a
  Metropolis-adjusted preconditioner, so spectral error enters the mixing constants and not the
  invariant target, and the full-cycle error obeys a recursion whose abstention term is governed by the
  geometric abstention load.

\end{enumerate}

The repair recovers eigenvalue laws, not eigenvectors, and this boundary separates the tasks the
tool serves from those it does not (Table~\ref{tab:scope}).

\begin{center}

\captionof{table}{What the repaired spectrum can and cannot support. The lift of Section~\ref{sec:7.1}
recovers a population eigenvalue law, not the population eigenbasis.}\label{tab:scope}

\begin{tabularx}{\linewidth}{@{}XX@{}}
\toprule
Depends on the eigenvalue law alone & Also depends on the eigenbasis \\
\midrule
spectral preconditioning; condition-number control; & subspace estimation; dimension reduction; \\
trace- and determinant-based detection statistics & modal decomposition \\
\bottomrule
\end{tabularx}
\end{center}

Throughout, \(d\) is the state dimension, \(N\) the ensemble size and \(c=d/N\).

Section~\ref{sec:3} states the forward problem and identifies the object to be conditioned on;
Section~\ref{sec:6} develops the inversion, its condition number and its certificate;
Section~\ref{sec:7} treats the target-preserving use of the recovered spectrum and the full cycle;
Section~\ref{sec:10} reports the pre-registered experiments. Proofs of the main results are in the
supplementary material.

\section{The forward problem and the conditioning object}\label{sec:3}

\subsection{Notation}\label{sec:3.1}

\textbf{Ensemble and second-moment matrices.} \(X=[x_1,\dots,x_N]\in\mathbb{R}^{d\times N}\); weights \(w_i\ge0\), \(\sum_iw_i=1\); \(\lambda_i=Nw_i\); \(\rho_N=\frac1N\sum_i\delta_{\lambda_i}\); \(\hat m=Xw\); \(S_w^u=\sum_iw_ix_ix_i^\top\); \(S_w^c=S_w^u-\hat m\hat m^\top\); counts \(n=(n_1,\dots,n_N)\) satisfying \textbf{(A1)} \(\sum_in_i=N\) a.s. and \textbf{(A2)} \(\mathbb{E}[n_i\mid X,w]=Nw_i\); \(\hat m_r=\frac1NXn\); \(S_r^u=\frac1N\sum_in_ix_ix_i^\top\); \(S_r^c=S_r^u-\hat m_r\hat m_r^\top\); \(k_i=\lfloor\lambda_i\rfloor\), \(u_i=\operatorname{frac}(\lambda_i)\), \(R_N=\sum_iu_i\). For a symmetric matrix \(A\), \(\mu_A\) denotes its empirical spectral distribution (ESD).

\textbf{Metrics.} \(d_{\rm BL}\) is the bounded-Lipschitz metric and \(W_1\) the
\(1\)-Wasserstein metric. The fixed-band Stieltjes discrepancy between finite measures \(\mu,\nu\)
on \(\mathbb{R}\), at bandwidth \(h>0\) over the segment \([0,\xi_{\rm hi}]\), is
\begin{equation}
d_{\rm St}(\mu,\nu):=d_{\rm St}^{(h,\xi_{\rm hi})}(\mu,\nu)
:=\sup\bigl\{|m_\mu(z)-m_\nu(z)|:\ z=\xi+ih,\ \xi\in[0,\xi_{\rm hi}]\bigr\},
\end{equation}
on the upper-half-plane branch \(\operatorname{Im}m_\mu>0\), with \(m_\mu(z)=\int(t-z)^{-1}\mu(dt)\).
Values on the segment determine \(\mu\), and weak convergence implies uniform convergence there,
so \(d_{\rm St}\) is a metric adapted to the limit theorems below. The pair \((h,\xi_{\rm hi})\) is
fixed in the theory (Remark~\ref{rem:2}); smaller \(h\) is more sensitive to local differences,
which is the bandwidth dependence of the activation zone (Section~\ref{sec:10.5}).

\textbf{Free probability.} \(\boxplus,\boxtimes\) are the free additive and
multiplicative convolutions \cite{ref09,ref26,ref27}. \(\mathcal{T}_c(\eta)\) is the covariance-side
spectral map, defined by \(c\,\mathcal{T}_c(\eta)+(1-c)\delta_0=D_c(\eta\boxtimes\mathrm{MP}(1/c))\)
with \(\mathrm{MP}\) the Marchenko--Pastur law \cite{ref28}, equivalently by the companion fixed-point
equation \(z=-1/m(z)+\int t(1+ctm(z))^{-1}\eta(dt)\). We write \(s_T(z):=\int(1-zx)^{-1}T(dx)\) for
the companion transform, \(\mathcal{L}_\mu(t):=\int e^{-tx}\mu(dx)\) for the Laplace transform, and
\(\nu_{\rm multi}(\rho):=\mathrm{MixPoi}(\rho)\), \(\nu_{\rm res}(\pi):=\mathcal{L}(K+\mathrm{Poi}(U))\)
with \((K,U)\sim\pi\) for the count laws.

\textbf{Resolution scale and classes.} \(\theta>0\) is the resolution scale of the spectral inversion (Section~\ref{sec:6.2}), a scalar distinct from the count law \(\nu\). \(\mathcal{H}_{a,b}:=\mathcal{P}([a,b])\) with \(0<a<b<\infty\), and \(\mathcal{U}\subseteq\mathcal{H}_{a,b}\) is the certified class, the relative neighborhood of \(H_0\) on which the condition number is evaluated; it is fixed exogenously and is never estimated from the data.

\subsection{High-dimensional and spectral assumptions}\label{sec:3.2}

\textbf{(G1)} \(d/N\to c\in(0,\infty)\). \textbf{(G2)} \(H_N=\mu_{\Sigma_N}\Rightarrow H\), \(\sup_N\|\Sigma_N\|_{\rm op}<\infty\), \(H\) compactly supported; \(a_1=\int\tau\,H(d\tau)\). \textbf{(G3)} \(Z_N=(z_{ij})\) has i.i.d. centered unit-variance entries; the main theorems use Gaussian entries, with universality left as an open direction. The fluctuation model is \(X_N=\Sigma_N^{1/2}Z_N\), with \(\operatorname{Var}(z_{ij})=1\); the sample-covariance normalization is carried by the factor \(1/N\) in \(S_r^u\), not by \(X_N\). This is the generalized sample-covariance setting of \cite{ref29}. Gaussianity is used at exactly one point: conditionally on the count diagonal, the whitened fluctuation matrix is orthogonally invariant and hence asymptotically free from the deterministic population factor \cite{ref09,ref17}.

\subsection{The resampling environment}\label{sec:3.4}

\textbf{(G4) (exogenous random environment)} There exist \(\mathcal{W}_N\) and auxiliary randomness \(U_N\) such that \((\mathcal{W}_N,U_N)\perp Z_N\); \(w_N=w_N(\mathcal{W}_N)\); and \(n_N=\mathsf{R}_N(w_N,U_N)\) for the measurable realization of the chosen scheme. With the extended environment \(\mathcal{E}_N:=\sigma(\mathcal{W}_N,n_N)\) we have \(\mathcal{E}_N\perp Z_N\), and conditionally on \(\mathcal{E}_N\) the diagonal \(D_N=\operatorname{diag}(n_N)\) is deterministic.

\textbf{(W2) (weight-law convergence)} \(\rho_N\Rightarrow\rho\) in probability and \(\frac1N\sum_i\lambda_i^2\to r_2:=\int\lambda^2\rho(d\lambda)<\infty\) in probability. Consequently \(\int\lambda\,\rho(d\lambda)=1\) and \(\max_i\lambda_i=O_p(\sqrt N)\).

\textbf{(M)/(R)} For multinomial resampling we assume (W2); for residual resampling we assume (W2) and \(\pi_N=\frac1N\sum_i\delta_{(k_i,u_i)}\Rightarrow\pi\) in probability.

\emph{Interpretation.} (G4) admits three legitimate sources: exogenous importance weights; sample
splitting or cross-fitting, where one subset of particles estimates the weight structure and an
independent subset carries the spectral ensemble; and conditionally exogenous local approximations. The
low-dimensional model \(w_i=\omega(B^\top x_i)\), \(q\ll d\), is treated as a proved proposition
rather than an assumption.

\subsection{Count laws under random weights}\label{sec:4}

\begin{lemma}[{Centering}]\label{lem:1}

Under (A1)--(A2), conditionally on \((X,w)\):

(i) \(\mathbb{E}[\hat m_r\mid X,w]=\hat m\) and \(\mathbb{E}[S_r^c\mid X,w]=S_w^c-B_n\preceq S_w^c\), where \(B_n=\frac1{N^2}XC_nX^\top\succeq0\) with \(C_n=\operatorname{Cov}(n\mid X,w)\). Resampling is unbiased for the uncentered second moment but downward biased for the centered covariance; the exact loss equals the conditional covariance of the resampled mean. Under \(\frac1N\sum_i\lambda_i\|x_i\|^2=O_p(d)\) one has \(\operatorname{tr}B_n=O_p(d/N)\) and \(d^{-1}\operatorname{tr}B_n=O_p(1/N)\). The matrix \(B_n\) may have rank up to \(d\); it is not low-rank.

(ii) \(S_w^c\preceq_{{\rm cx}\mid X,w}S_r^c+B_n\): for every convex \(\Phi\) with existing expectations, \(\Phi(S_w^c)\le\mathbb{E}[\Phi(S_r^c+B_n)\mid X,w]\).

(iii) (Finite-rank invariance of the ESD) \(\operatorname{rank}(S_r^u-S_r^c)\le1\), hence \(\sup_x|F^{S_r^c}(x)-F^{S_r^u}(x)|\le1/d\); as \(d\to\infty\) the centered and uncentered laws coincide and the zero-atom masses differ by at most \(1/d\).

(iv) (Boundary note) This concerns only the ESD and normalized nullity: extreme eigenvalues, outliers, and unbounded spectral moments are not preserved by finite-rank perturbations. Edge convergence transfers only under an additional no-outlier condition with adjacent extreme order statistics sharing a limit.

\end{lemma}

\emph{Proof.} See Section~S-A of the supplementary material.

\begin{remark}[{Why the conditioning object is $\mathcal{E}_N$ and not $\mathcal{W}_N$}]\label{rem:1}

The distinction matters already at the level of the count laws. The multinomial count vector is a function of the weights \emph{and} of fresh resampling randomness, and it is this joint object that enters the resampled second-moment matrix. Conditioning on \(\mathcal{W}_N\) therefore leaves the count diagonal random, whereas conditioning on \(\mathcal{E}_N=\sigma(\mathcal{W}_N,n_N)\) freezes it. Since the deterministic spectral theorem requires a deterministic diagonal, \(\mathcal{E}_N\) is the minimal conditioning under which that theorem applies, and (G4) is exactly the requirement that this conditioning is compatible with the fluctuation matrix, \(\mathcal{E}_N\perp Z_N\).

\end{remark}

\begin{theorem}[{Count laws under random weights}]\label{thm:1}

Assume (W2) (and \(\pi_N\Rightarrow\pi\) in the residual case). Then, in probability:

(i) (Multinomial) \(\nu_N=\frac1N\sum_i\delta_{n_i}\Rightarrow\mathrm{MixPoi}(\rho)\) in \(d_{\rm BL}\);

(ii) (Residual) \(\nu_N\Rightarrow\mathcal{L}(K+\mathrm{Poi}(U))\), \((K,U)\sim\pi\);

(iii) (Second count moments) \(\frac1N\sum_in_i^2\to1+r_2\) (multinomial), resp. \(r_2+\mathbb{E}_\pi U\) (residual).

\end{theorem}

The proof uses a balls-into-bins representation with an Efron--Stein variance bound \cite{ref31} for bounded test functions, Le Cam Poissonization \cite{ref32} for expectations, and exact factorial-moment computations for the unbounded second-moment statistic.

\emph{Proof.} See Section~S-B of the supplementary material.

\subsection{The random-environment spectral law}\label{sec:5}

\begin{theorem}[{Deterministic core used in the proof of \cite[Theorem 12]{ref01}}]\label{thm:2}

For deterministic diagonal \(\widetilde D_N\) with entries in \(\mathbb{Z}_{\ge0}\), ESD \(\Rightarrow\widetilde\eta\) (probability law, mean \(\le1\)), \(\frac1N\sum_i\widetilde d_i^{\,2}\le K<\infty\), Gaussian \(Z_N\) independent of \(\widetilde D_N\), and \(\Sigma_N\) satisfying (G2), the ESD of \(\frac1N\Sigma_N^{1/2}Z_N\widetilde D_NZ_N^\top\Sigma_N^{1/2}\) converges in probability to \(H\boxtimes\mathcal{T}_c(\widetilde\eta)\).

\end{theorem}

This deterministic-diagonal statement is the imported step used in the proof of
\cite[Theorem 12]{ref01}, verified in \cite[SM \S{}S7]{ref01}. Its deterministic-diagonal form,
rather than a count-law form, is what makes the conditioning argument work.

\begin{lemma}[{Continuity of the spectral map}]\label{lem:2}

If \(\eta_L\Rightarrow\eta\) with \(\int t\,\eta_L(dt)\) uniformly bounded, then \(\mathcal{T}_c(\eta_L)\Rightarrow\mathcal{T}_c(\eta)\).

\end{lemma}

\emph{Proof.} See Section~S-C of the supplementary material.

\begin{theorem}[{Random-environment spectral law}]\label{thm:3}

Under (G1)--(G4) with (M) or (R), with \(\nu=\nu_{\rm multi}(\rho)\) resp. \(\nu_{\rm res}(\pi)\),
\begin{equation}\mu_{S_r^u}\Rightarrow H\boxtimes\mathcal{T}_c(\nu)\quad\text{in probability.}\end{equation}
\end{theorem}
\emph{Proof.} See Section~S-D of the supplementary material.

\begin{corollary}[{Centered law}]\label{cor:1}

Under the same assumptions, \(\mu_{S_r^c}\Rightarrow H\boxtimes\mathcal{T}_c(\nu)\) in probability. Centering does not re-enter the random-matrix argument; it is a rank-one ESD perturbation (Lemma~\ref{lem:1}(iii)).

\end{corollary}

\begin{corollary}[{Closed-form reference quantities}]\label{cor:2}

Let \(\mu=H\boxtimes\mathcal{T}_c(\nu)\). Then, with all four items taken from \cite{ref01} except
where noted:

\begin{enumerate}

\item

  \textbf{Zero atom.} \(\mu(\{0\})=\max\{H(\{0\}),\mathcal{T}_c(\nu)(\{0\})\}\), and under multinomial resampling \(\mathcal{T}_c(\nu)(\{0\})=\max(0,1-(1-\nu(\{0\}))/c)\).

\item

  \textbf{Active-column fraction.} \(p_s:=1-\nu(\{0\})\); under multinomial resampling \(p_s=1-\mathcal{L}_\rho(1)\). The rank threshold is \(p_s\gtrless c\); \(p_s\) alone does not determine the general spectral gap.

\item

  \textbf{Rank-only benchmark.} If \(H=\delta_1\) and \(D_N\) is asymptotically an active-column projection, the Marchenko--Pastur benchmark is \((\sqrt c-\sqrt{p_s})^2\); it is not a universal lower-edge bound for general \((H,\nu)\).

\item

  \textbf{Second moment.} If \(m_1(\nu)=1\), then \(m_2(\mu)=m_2(H)+c\,m_1(H)^2m_2(\nu)\), with \(m_2(\nu)=1+r_2\) (multinomial) resp. \(r_2+\mathbb{E}_\pi[U]\) (residual); the random-weight instantiation is Theorem~\ref{thm:1}(iii).

\end{enumerate}
\end{corollary}

\begin{lemma}[{Weight-only diagnostics cannot separate spectra}]\label{lem:weight-only}

Fix a weight vector \(w\) and the resampling scheme, so that the distortion law
\(T=\mathcal{T}_c(\nu)\) is fixed, and let \(\mathcal{D}=\psi(w)\) be any diagnostic that is a
function of the weights alone. If the forward map is injective on the candidate class, there exist
\(H_1\neq H_2\) on which \(\mathcal{D}\) takes the same value while the limiting laws
\(H_1\boxtimes T\) and \(H_2\boxtimes T\) differ.

\end{lemma}

\emph{Proof.} Both data sets share \(w\), hence \(\nu\), \(T\) and the value of \(\psi\); by
injectivity on the class (Lemma~\ref{lem:3}(iii)) the two limiting laws differ. \hfill\(\square\)

The statement is deliberately confined to the class of functionals the filter has access to
\emph{before} it looks at the spectrum. It is not a statement about scalar diagnostics in general: the
zero-atom mass is itself a scalar and separates some spectra, and on compact support a moment sequence
determines the measure. It is also not a statement of dynamical blindness; the monitor experiment of
Section~\ref{sec:10.6} shows the ESS and the zero-atom mass moving in synchrony. What fails is the
identification of the distortion law, and with it the possibility of directing a repair.

\begin{remark}[{The zero atom as lost second-order dimension}]\label{rem:dim}

Let \(m_0:=\mu(\{0\})\) be the zero-atom mass of the limiting law \(\mu=H\boxtimes\mathcal{T}_c(\nu)\),
given in closed form by Corollary~\ref{cor:2}(1). Asymptotically the empirical second-moment matrix
carries nonzero mass on a fraction \(1-m_0\) of the spectrum, so the mass on the atom contributes
nothing to a spectral preconditioner, to a condition number, or to a trace-based detection statistic.
The quantity \((1-m_0)d\) is therefore an effective-dimension proxy for spectral preconditioning and
condition-number control, not a statement of exact finite-\(N\) rank or of an invariant subspace.
This is a statement about the limiting law, not about the rank of any finite-\(N\) matrix. Weight-only
summaries can supply partial population statistics of the weights but cannot determine the full
distortion law or its zero-atom mass, so they cannot identify the distortion or direct a repair; nor
is this a claim of dynamic blindness, since the monitor experiment of Section~\ref{sec:10.5} shows
that the two can move in synchrony.

\end{remark}

One consequence is used later: with \(a_1=m_1(H)\) and \(m_1(\nu)=1\), residual resampling never exceeds multinomial in the second spectral moment, \(m_{2,\rm residual}=m_{2,H}+ca_1^2(r_2+\mathbb{E}_\pi[U])\le m_{2,\rm multi}\), and the active-column fraction is nonincreasing in the convex order (Section~S2 of the supplementary material). These four items are the reference values against which the experiments of Section~\ref{sec:10} are checked.

\section{Inversion and certification}\label{sec:6}

\subsection{The inversion problem and the condition number}\label{sec:6.1}

The observed object is \(\widehat\mu_N=\mu_{S_r^c}\), and Corollary~\ref{cor:1} gives
\(\varepsilon_N:=d_{\rm St}(\widehat\mu_N,\mu)\to0\) in probability with \(\mu=H_0\boxtimes T\), where
the distortion law \(T=\mathcal{T}_c(\nu)\) is computable from the weights. The task is to recover
\(H_0\) from \(\widehat\mu_N\) and the weights alone. The candidate class is
\(\mathcal{H}_{a,b}=\mathcal{P}([a,b])\), \(0<a<b<\infty\), compact and away from the zero-atom and
hard-edge regions where inversion degenerates, and the forward operator is
\(\mathcal{F}_T(G):=G\boxtimes T\).

\subsection{Structure of the forward map}\label{sec:6.2}

\begin{lemma}[{Structure of the forward map}]\label{lem:3}

(i) \(\mathcal{H}_{a,b}\) is weakly compact. (ii) \(G\mapsto G\boxtimes T\) is weakly continuous on \(\mathcal{P}([a,b])\). (iii) Let \(T\in\mathcal{P}(\mathbb{R}_{\ge0})\) satisfy \(0<m_1(T)<\infty\). Then \(\mathcal{F}_T\) is injective on \(\mathcal{H}_{a,b}\): \(S_{G\boxtimes T}=S_GS_T\), and \(S_T\) is defined and nonzero on a left interval \((-\delta,0]\), so division recovers \(S_G\); compactly supported measures are moment-determinate.

\end{lemma}

\emph{Proof.} See Section~S-E of the supplementary material.

\textbf{Resolution-restricted condition number.} Injectivity does not provide a uniform Lipschitz inverse. Let \(\mathcal{U}\subseteq\mathcal{H}_{a,b}\) be the certified class and let \(\theta>0\) be a resolution scale. Write
\begin{equation}
\kappa_{\rm inv}^{(\theta)}(H_0,T;\mathcal{U}):=\inf_{\substack{G_1\neq G_2\in\mathcal{U}\\ W_1(G_1,G_2)\ge\theta}}\frac{d_{\rm St}(\mathcal{F}_T(G_1),\mathcal{F}_T(G_2))}{W_1(G_1,G_2)} .
\end{equation}

The restriction \(W_1\ge\theta\) is essential: without it the ratio has no positive lower bound
in general, since a homeomorphism that contracts measure differences near the diagonal drives the
infimum to zero (\(F(x)=x^2\) on \([0,1]\), where \(|x^2-y^2|/|x-y|=x+y\)), so the unrestricted
quantity cannot carry an error budget. With the restriction it is finite and strictly positive by
Theorem~\ref{thm:6}. We work with \(d_{\mathcal{H}}:=W_1\) throughout, so the inversion budget is a
\(W_1\) budget and connects to the cycle analysis of Section~\ref{sec:8}: a mass-\(t\) atom moved over
a fixed distance costs \(O(t)\) in both \(W_1\) and the fixed-bandwidth Stieltjes discrepancy, which is
what makes the two commensurable.

\begin{remark}[{Implemented discrepancy}]\label{rem:2}
The registered implementation (Section~\ref{sec:10.1}) uses a root-mean-square discretization of
$d_{\rm St}$ over $n_x$ grid points, with the zero atom removed (which leaves the condition number
unchanged by the maximum-atom rule) and a data-dependent band converging to a fixed band; the vanishing
of $\varepsilon_N,\delta_N$ and Theorem~\ref{thm:5} hold verbatim for it. On the grid-supported class
used in the computations, identifiability additionally requires injectivity of
$G\mapsto(m^+_{\mathcal{F}_T(G)}(z_j))_{j\le n_x}$, where $m^+_\mu(z):=m_\mu(z)+\mu(\{0\})/z$ denotes the Stieltjes transform with the zero atom removed, checked locally by $\widehat\kappa_{\rm inv}>0$,
and the implemented budget $\widehat{\mathcal{B}}_N$ is a consistent estimate, not a certified bound
(Section~P1 of the supplementary material).
\end{remark}

\subsection{The plug-in distortion law}\label{sec:6.3}

We do not estimate the limiting law \(\rho\); the plug-in distortion law is built directly from the realized weights:
\begin{equation}
\widehat\nu_N^{\rm multi}:=\frac1N\sum_{i=1}^N\mathrm{Poi}(\lambda_i),\qquad \widehat\nu_N^{\rm res}:=\frac1N\sum_{i=1}^N\mathcal{L}(k_i+\mathrm{Poi}(u_i)), \qquad \widehat T_N:=\mathcal{T}_{c_N}(\widehat\nu_N),\quad c_N=d/N . 
\end{equation}
\begin{lemma}[{Plug-in lemma}]\label{lem:4}

(i) Multinomial: \(\rho_N\Rightarrow\rho\) under (W2) implies \(\widehat\nu_N^{\rm multi}\Rightarrow\mathrm{MixPoi}(\rho)\); (ii) residual: \(\pi_N\Rightarrow\pi\) implies \(\widehat\nu_N^{\rm res}\Rightarrow\mathcal{L}(K+\mathrm{Poi}(U))\); (iii) in both cases \(m_1(\widehat\nu_N)=1\) exactly and \(m_2(\widehat\nu_N)=O_p(1)\), so Lemma~\ref{lem:2} gives \(\widehat T_N\Rightarrow T\); (iv) joint continuity of \((G,Q)\mapsto G\boxtimes Q\) on compact sets gives \(\delta_N:=\sup_{G\in\mathcal{H}_{a,b}}d_{\rm St}(G\boxtimes\widehat T_N,G\boxtimes T)\to0\) in probability.

\end{lemma}

\emph{Proof.} See Section~S-E of the supplementary material.

Of the two available laws, \(\widehat T^{\rm cond}\) and
\(\widehat T^{\rm realized}:=\mathcal{T}_{c_N}(\nu_N)\), the realized-count law inverts better at the
largest ensemble size of Section~\ref{sec:10.2} and the conditional plug-in is smoother, which the
mismatch diagnostic requires of its reference law; we retain \(\widehat T^{\rm cond}\) as default and
claim no strict ordering.

\subsection{The estimator and its consistency}\label{sec:6.4}
\begin{equation}
 \widehat H_N\in\arg\min_{G\in\mathcal{H}_{a,b}}\Bigl\{d_{\rm St}\bigl(\widehat\mu_N,\mathcal{F}_{\widehat T_N}(G)\bigr)^2+\gamma_NR(G)\Bigr\}, 
\end{equation}
a Tikhonov-regularized minimum-distance problem \cite{ref44} with \(R\) lower semicontinuous; an
argmin exists by lower semicontinuity on a compact set. Quadratic penalties \(R(G)=\int x^2G(dx)\)
are safe defaults, entropy penalties being typically infinite on atomic laws. The schedule must be
scale-free: a fixed \(\gamma_N=\gamma_0/N\) does not adapt to the scale of \(H\), because the penalty
enters through \(R(H_0)\), and in the strong-anisotropy regime can displace the spike mass entirely,
so we use \(\gamma_N=\gamma_{\rm rel}/\widehat R_N\).

\begin{theorem}[{Consistency and local error budget}]\label{thm:4}

Assume (i) \(H_0\in\mathcal{H}_{a,b}\) and the ideal model \(\mu=\mathcal{F}_T(H_0)\); (ii) \(R(H_0)<\infty\); (iii) \(\varepsilon_N\to0\) in probability (Corollary~\ref{cor:1}); (iv) \(\delta_N\to0\) in probability (Lemma~\ref{lem:4}; in the oracle setting \(\delta_N=0\)); (v) \(\gamma_N\to0\). Then \(W_1(\widehat H_N,H_0)\to0\) in probability. Moreover, with \(\kappa:=\kappa_{\rm inv}^{(\theta)}(H_0,T;\mathcal{U})\), which is strictly positive by Theorem~\ref{thm:6}, and \(\mathcal{B}_N:=\kappa^{-1}[\delta_N+\sqrt{(\varepsilon_N+\delta_N)^2+\gamma_NR(H_0)}+\varepsilon_N]\), on the event \(\{\widehat H_N\in\mathcal{U}\}\), which has probability tending to one,
\begin{equation}
W_1(\widehat H_N,H_0)\le\max\bigl\{\theta,\ \mathcal{B}_N\bigr\}.
\end{equation}

\end{theorem}

Consistency is proved without the condition number, from compactness and injectivity; the budget is the quantitative refinement, informative only at scale $\mathcal{B}_N\ge\theta$, and below $\theta$ the max is trivial. Throughout, $\mathcal{B}_N$ is read at fixed $\theta$ with the maximum suppressed; since $\mathcal{B}_N=o_p(1)$ under (iii)--(v), consistency statements are unaffected.

\emph{Proof.} See Section~S-F of the supplementary material.

\begin{theorem}[{Consistent mismatch diagnostic}]\label{thm:5}

Suppose \(\mathcal{H}_{a,b}\) is compact, \(\varepsilon_N\to0\) and \(\delta_N\to0\) in probability, and define \(\varepsilon_{\rm model}:=\inf_{G\in\mathcal{H}_{a,b}}d_{\rm St}(\mu,G\boxtimes T)\) and its computable proxy \(\widehat\varepsilon_{{\rm model},N}:=\inf_{G\in\mathcal{H}_{a,b}}d_{\rm St}(\widehat\mu_N,G\boxtimes\widehat T_N)\). Then \(|\widehat\varepsilon_{{\rm model},N}-\varepsilon_{\rm model}|\le\varepsilon_N+\delta_N\to0\) in probability.

\end{theorem}

\emph{Proof.} See Section~S-G of the supplementary material. The theoretical certificate is the global minimum-distance residual; an implemented certificate additionally carries a numerical optimization error \(\eta_N^{\rm opt}\).

\emph{Model mismatch.} If \(\mu\notin\mathcal{F}_T(\mathcal{H}_{a,b})\), the estimator targets the
pseudo-true spectrum \(H^\dagger\in\arg\min_{G\in\mathcal{H}_{a,b}}d_{\rm St}(\mu,\mathcal{F}_T(G))\), and
\(\widehat\varepsilon_{\rm model}\) certifies the forward residual, not \(W_1(\widehat H_N,H_0)\). If
\(H^\dagger\) is unique with \(\kappa_{\rm inv}^{(\theta)}(H^\dagger,T;\mathcal{U})>0\), the same
argument gives \(W_1(\widehat H_N,H^\dagger)\to0\) at the local rate, with \(W_1(H^\dagger,H_0)\) the
irreducible spectral bias of misspecification.

\begin{theorem}[{Identifiability: strict positivity at finite resolution}]\label{thm:6}

Let \(\mathcal{U}\subseteq\mathcal{H}_{a,b}\) be closed. Under Lemma~\ref{lem:3}, \(\mathcal{F}_T\) is a homeomorphism of \(\mathcal{U}\) onto its image, so its inverse is uniformly continuous: for every resolution scale \(\theta>0\) there exists \(\delta_\theta>0\) such that
\begin{equation}
W_1(G_1,G_2)\ge\theta\quad\Longrightarrow\quad d_{\rm St}\bigl(\mathcal{F}_T(G_1),\mathcal{F}_T(G_2)\bigr)\ge\delta_\theta .
\end{equation}

Since \(W_1\le b-a\) on \(\mathcal{H}_{a,b}\), the resolution-restricted condition number of Section~\ref{sec:6.2} satisfies
\begin{equation}
\kappa_{\rm inv}^{(\theta)}(H_0,T;\mathcal{U})\ \ge\ \frac{\delta_\theta}{b-a}\ >\ 0 . 
\end{equation}
\end{theorem}

The modulus $\delta_\theta$ is not explicit, and the restriction is intrinsic: two measures on a common compact support can have arbitrarily close transforms while remaining at a fixed $W_1$ distance, so no support-free explicit constant exists and none is claimed. What degenerates as $\theta\downarrow0$ is the modulus, not the positivity, so the degradation measured in Section~\ref{sec:10.3} concerns conditioning rather than existence. The theorem is a positivity statement, not a computable region; it supplies what the budget of Theorem~\ref{thm:4} requires, and the modulus is assessed numerically on the two-point family (Section~\ref{sec:10.3} and Section~S-K of the supplementary material), no constant from which enters any theorem. Per scheme, $T=\mathcal{T}_c(\nu_{\rm multi}(\rho))$ or $T=\mathcal{T}_c(\nu_{\rm res}(\pi))$, the residual case requiring $\pi$ because the floor--fraction map is discontinuous at integers.

\begin{remark}[{Conditioning mechanism}]\label{rem:3}
The quantity that tracks the condition number is the zero-atom mass $q$ of the distortion law, not the distance $|c-p_s|$ to the onset boundary, where the atom is born with zero mass and inversion is well conditioned (Section~\ref{sec:10.3}). With $m_1(T)$ pinned at one, Cauchy--Schwarz gives $m_2(T)\ge\bigl(\int_{x>0}x\,T(dx)\bigr)^2/(1-q)=(1-q)^{-1}$, so a growing atom forces a more dispersed $T$, which contracts the Stieltjes discrepancy relative to $W_1$: the inequality direction that obstructs inversion, and an instance of the smoothing effect of free convolution \cite{ref10}. This is a consistency argument, not a rate statement; Theorem~\ref{thm:6} guarantees positivity.
\end{remark}

\section{Target-preserving use and the full cycle}\label{sec:7}

\subsection{Target-preserving rejuvenation}\label{sec:7.1}

The recovered spectrum is only ever used as a proposal preconditioner, never as a target. Write
\(\mathcal{F}_{k-1}^N=\sigma(\text{particle system and all randomness up to }k-1)\) and let the
computable target at step \(k\) be the \(\mathcal{F}_{k-1}^N\)-conditional measure
\begin{equation}
\widetilde\pi_k^N(dx):=\frac{g_k(y_k\mid x)\,\widehat\nu_{k\mid k-1}^N(dx)}{\widehat\nu_{k\mid k-1}^N(g_k(y_k\mid\cdot))},
\qquad
\widehat\nu_{k\mid k-1}^N(dx)=\frac1N\sum_{j=1}^Np_k(dx\mid x_{k-1}^j).
\end{equation}
The rejuvenation kernel \(R_k\) consists of \(L_k\) random-walk Metropolis--Hastings (MH) steps
\cite{ref45} with target \(\widehat\pi_k^N=\widetilde\pi_k^N\), evaluated in density form. Because the
target is the computable SMC target and not a surrogate built from the recovered spectrum, that
spectrum enters only through the proposal. A surrogate target is not used: spectral-law error is not
state-space target error, and controlling the former would require a target-approximation theorem that
Theorems~\ref{thm:4} and~\ref{thm:5} do not supply.

The proposal uses \(M_k=\widehat\Sigma_k^{\rm lift}+\varepsilon I\) with
\(\widehat\Sigma_k^{\rm lift}=U_k\operatorname{diag}(Q_{\widehat H_{k,\gamma}}((j-\tfrac12)/d))_{j=1}^dU_k^\top\),
where \(Q_G\) is the quantile function of \(G\), \(U_k\) is the ordered eigenvector matrix of
\(S_{r,k}^c\) and the quantiles follow the nondecreasing eigenvalue order. The lift is rotationally
equivariant and recovers a population eigenvalue law, not the eigenbasis; the \(\varepsilon\)-floor
prevents a degenerate proposal and is not a geometric-ergodicity condition, its value being set by the
posterior anisotropy \(\kappa(C_\pi)=s_{\max}(s_{\min}+r)/[s_{\min}(s_{\max}+r)]\), capped at
\(1+r/s_{\min}\) under Gaussian linear observations, and by the usability of the empirical eigenbasis,
which degrades at high \(c\) with a macroscopic zero atom. The damped Gaussian-proxy transport
\(G_\beta\), of closed-form cost \cite{ref18}, is only a proposal mechanism, with the full Hastings
correction whenever it makes the proposal asymmetric.

\subsection{Invariance and the error decomposition}\label{sec:7.2}

\begin{lemma}[{Target preservation}]\label{lem:6}

The MH kernel is reversible for its actual target, \(\widehat\pi_k^NR_k=\widehat\pi_k^N\), by
construction. Since the actual target is the computable SMC target,
\(\widetilde\pi_k^NR_k=\widetilde\pi_k^N\): the move step preserves the conditional SMC target and
introduces no target bias.

\end{lemma}

\begin{theorem}[{Conditional error decomposition}]\label{thm:7}

Work conditionally on \(\mathcal{F}_{k-1}^N\). If the MH kernel satisfies a drift--minorization
condition with \(\mathcal{F}_{k-1}^N\)-measurable constants \(C_k,\varrho_k\), then in a metric
compatible with the mixing theorem
\(d\bigl(\mathcal{L}(X_{k,L_k}^{\rm alg}\mid\mathcal{F}_{k-1}^N),\widetilde\pi_k^N\bigr)\le C_k\varrho_k^{L_k}\),
and hence \(d(\pi_k,\mathcal{L}(X_{k,L_k}^{\rm alg}\mid\mathcal{F}_{k-1}^N))\le d(\pi_k,\widetilde\pi_k^N)+C_k\varrho_k^{L_k}\).
For each fixed \(k\), Feynman--Kac stability gives \(d_{\rm BL}(\widetilde\pi_k^N,\pi_k)\to0\) in
probability independently of the number of move steps, and the mixing term is made small by raising
\(L_k\) at no asymptotic bias. The recovered spectrum affects only the proposal, hence only
\(C_k,\varrho_k\), not the invariant target.

\end{theorem}

\emph{Proof.} See Section~S-H of the supplementary material. The numerical validation of
Section~\ref{sec:10} covers random-walk Metropolis--Hastings; the Metropolis-adjusted Langevin
algorithm lies within the scope of Theorem~\ref{thm:7} without numerical validation.

\subsection{The full-cycle recursion under scalar stable dynamics}\label{sec:8}

\emph{The one-cycle operator.} One cycle maps \(H_k^+\mapsto H_{k+1}^+\) in three stages: propagation, which is not spectrally closed
for general dynamics because the spectrum of \(A\Sigma A^\top+Q\) depends on eigenvector orientation;
selection, \(H_{k+1}^-\to H_{k+1}^-\boxtimes\mathcal{T}_c(\nu_{k+1})\) (Theorem~\ref{thm:3}); and
correction, with \(W_1(\widehat H_{k+1},H_{k+1}^-)\le\mathcal{B}_{k+1}\) (Theorem~\ref{thm:4}),
setting \(H_{k+1}^+:=\widehat H_{k+1}\). Fig.~\ref{fig:1} summarizes the corrected cycle.

\emph{Scope note.} The ideal object is the propagation spectrum with resampling distortion removed; no Bayesian posterior contraction is involved and end-to-end filtering accuracy remains outside the claims. The per-step budget carries the resolution ceiling of Section~\ref{sec:6.4}, written \(\mathcal{B}_k^{\theta}:=\max\{\theta,\mathcal{B}_k\}\) where the ceiling is visible.

\textbf{Setting.} \(x_{k+1}=a_kx_k+\xi_k\), \(\operatorname{Cov}(\xi_k)=q_kI\), \(|a_k|\le\bar a<1\), \(0<q_{\min}\le q_k\le q_{\max}\). Propagation is spectrally exact with no freeness assumption: \(\Sigma_{k+1}^-=a_k^2\Sigma_k+q_kI\) shares the eigenbasis of \(\Sigma_k\), so \(H_{k+1}^-=(D_{a_k^2})_\#H_k^+\boxplus\delta_{q_k}\). The corrected cycle is summarized in Fig.~\ref{fig:1}.

\begin{center}
\includegraphics[width=0.90\linewidth]{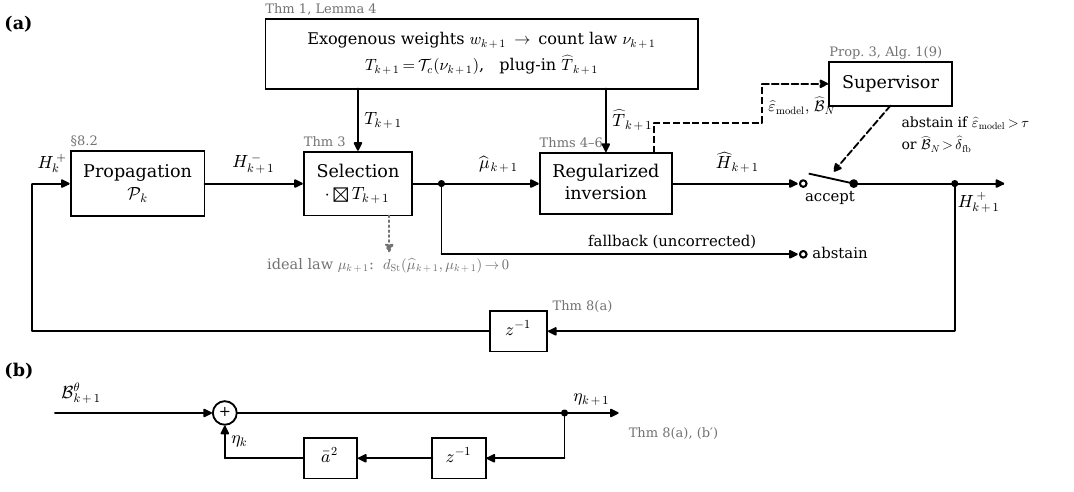}
\captionof{figure}{The spectrally corrected filtering cycle and its error recursion. (a) One cycle: propagation, selection by resampling (Theorem~\ref{thm:3}), and regularized inversion (Theorems~\ref{thm:4}--\ref{thm:6}); the supervisor accepts the repair or abstains (Proposition~\ref{prop:3}). Dashed arrows carry diagnostics and the control signal. (b) The comparison system majorizing $\eta_k=W_1(H_k^+,H_k^{\star,+})$: the recursion $\eta_{k+1}\le\bar a^2\eta_k+\mathcal{B}^\theta_{k+1}$ has pole $\bar a^2<1$ (Theorem~\ref{thm:8}(a)); it depicts an upper bound, not the error dynamics.}\label{fig:1}
\end{center}

\begin{theorem}[{Full-cycle recursion}]\label{thm:8}

Assume (i) scalar dynamics as above; (ii) (W2) at each step with \(\sup_{k\le K}r_2(\rho_k)<\infty\), the time-uniform statements additionally requiring the envelope of (b$^{\prime}$); (iii) correction at each step within a uniformly identifiable region, \(\kappa_{{\rm inv},k}^{(\theta)}\ge\kappa_0>0\) at a fixed \(\theta\); (iv) (G1)--(G4) at each step; (v) a common interval \([a,b]\) positively invariant under the recursion, which holds when \(a\le q_{\min}\) and \(b\ge\max\{\sup\operatorname{supp}(H_0^+),q_{\max}/(1-\bar a^2)\}\).

Define the ideal laws \(H_k^{\star,+}\) by \(H_{k+1}^{\star,-}=(D_{a_k^2})_\#H_k^{\star,+}\boxplus\delta_{q_k}\) together with \(H_k^{\star,+}:=H_k^{\star,-}\), and set \(\eta_k:=W_1(H_k^+,H_k^{\star,+})\).

(a) \emph{Error recursion}: \(\eta_{k+1}\le\bar a^2\eta_k+\mathcal{B}_{k+1}^{\theta}\), using \\
\(W_1((D_t)_\#\mu,(D_t)_\#\nu)=tW_1(\mu,\nu)\) and the shift isometry.

(b) \emph{Fixed horizon}: for any fixed \(K\), as \(\min_{k\le K}N_k\to\infty\) one has \(\mathcal{B}_{k+1}=o_p(1)\) uniformly for \(k\le K\) (this statement concerns the uncapped budget; the caps \(\mathcal{B}_{k+1}^{\theta}\) used in the recursion below are bounded below by \(\theta\) and do not tend to zero), and \(\eta_K\le\bar a^{2K}\eta_0+\sum_{j<K}\bar a^{2(K-1-j)}\mathcal{B}_{j+1}^{\theta}\).

(b$^{\prime}$) \emph{Long horizon}: if a time-uniform envelope \(\mathcal{B}_k\le b_k\) holds simultaneously with probability tending to one and \(\sup_kb_k<\infty\), then

\begin{equation}\limsup_k\eta_k\ \le\ \frac{\max\{\theta,\ \sup_k b_k\}}{1-\bar a^2} .\end{equation}

If moreover \(b_k\to0\), the bound reduces to \(\theta/(1-\bar a^2)\). This is a floor of the certified bound, not of the error: as \(N_k\) increases the certificate descends to the resolution scale at which the conditioning statement holds and no further, while the error itself may be smaller. The floor is intrinsic, since $\theta\downarrow0$ leaves the modulus $\delta_\theta$ unbounded below and the budget diverges; \(\theta\) cannot be recovered from the data. A per-time \(o_p(1)\) statement does not yield time-uniform control, so the two horizons are stated separately.

\end{theorem}

\emph{Proof.} See Section~S-I of the supplementary material.

\subsection{The cost of abstention}\label{sec:8.3}

\begin{proposition}[{Cost of abstention}]\label{prop:3}

In the setting of Theorem~\ref{thm:8}, let \(\mathcal{A}\) be the supervisor's abstention set and define the geometric abstention load \(\Lambda_K:=\sum_{j\in\mathcal{A}\cap[1,K]}\bar a^{2(K-j)}\). Then
\begin{equation}
\eta_K\le\bar a^{2K}\eta_0+\sum_{j=1}^K\bar a^{2(K-j)}\bigl[b_j\mathbf{1}_{\{j\notin\mathcal{A}\}}+\delta_{{\rm fb},j}\mathbf{1}_{\{j\in\mathcal{A}\}}\bigr],
\end{equation}

where \(b_j\) and \(\delta_{{\rm fb},j}\) are the per-step budgets of corrected and fallback
steps, each carrying the resolution ceiling of Section~\ref{sec:6.4}. Consequently:
(i) \(\mathcal{A}=\varnothing\) recovers Theorem~\ref{thm:8}. (ii) If \(\sup_kb_k\le\bar b\) and
\(\sup_k\delta_{{\rm fb},k}\le\bar\delta\), then with both budgets capped and
\(\bar b^{\theta}:=\max\{\theta,\bar b\}\), \(\bar\delta^{\theta}:=\max\{\theta,\bar\delta\}\),
\[
\limsup_K\eta_K\le\frac{\bar b^{\theta}}{1-\bar a^2}+(\bar\delta^{\theta}-\bar b^{\theta})\limsup_K\Lambda_K,
\qquad
\Lambda_K\le\frac{1-\bar a^{2A_K}}{1-\bar a^2}.
\]
(iii) If \(b_k\to0\) and \(\Lambda_K\to0\) then \(\limsup_K\eta_K\le\theta/(1-\bar a^2)\), the
resolution floor of Theorem~\ref{thm:8}(b$^{\prime}$): the operative condition for \emph{approaching}
that floor is vanishing geometric load, not merely \(A_K=o(K)\), because abstentions clustered at the
end of the horizon are fully felt, while the floor itself is not removed by any abstention schedule.
(iv) If \(H_k^{\star,+}\notin\mathcal{H}_{a,b}\), every algorithm constrained to \(\mathcal{H}_{a,b}\)
incurs error at least \(\operatorname{dist}_{W_1}(H_k^{\star,+},\mathcal{H}_{a,b})\) at that step, so
abstention is necessary for any algorithm so constrained whenever this distance exceeds
\(\delta_{{\rm fb},k}\): the abstention rule is not conservatism but a feasibility constraint of the
candidate class.

\end{proposition}

\emph{Proof.} See Section~S-I of the supplementary material.

Without correction each selection injects a further multiplicative factor: along successive pure
distortion segments the zero mass obeys the maximum-atom rule rather than additive accumulation, and
intervening propagation may alter the relation, so no closed long-time formula is claimed. The
corrected recursion instead resets the spectrum to a \(\mathcal{B}_k^{\theta}\)-neighbourhood of
\(H^-\) at each step; the contrast between compounding and resetting is the comparative hypothesis of
the experiments of Section~\ref{sec:10.4}.

\subsection{The corrected filtering step}\label{sec:9.1}

The observation is absorbed exactly once, at the weight update; every later stage consumes information
only through weights, spectra, or closed-form diagnostics (Fig.~\ref{fig:1}). A trigger can
short-circuit the correction and the supervisor can abstain once the diagnostics are measured, every
trigger and abstention being logged by a reason code. The complete step-by-step specification is
given in Section~P3 of the supplementary material. The
per-step cost is \(O(d^3+Nd^2)\), the same order as an ensemble Kalman step and without Sinkhorn-type
\(O(N^2)\) cost; the forward-map solver passes seven unit tests
on the upper-half-plane branch, including a fixed-point residual below \(3\times10^{-13}\)
(Section~P1 of the supplementary material). The recovered spectrum enters only through the preconditioner, and the lift is used only
where the empirical eigenbasis is usable and the posterior anisotropy nontrivial; outside that regime
the identity preconditioner is competitive or better and the abstention rule decides.

\section{Numerical experiments}\label{sec:10}

\subsection{Pre-registration and provenance protocol}\label{sec:10.1}

Every figure and table below answers one pre-registered question: targets, sample sizes, repetition
counts and acceptance criteria were frozen before the runs, and every printed number is traceable
through a four-level chain in the deviation ledger. The discrepancy follows Remark~\ref{rem:2} with
\(h=0.03\,\xi_{\rm hi}\) and \(n_x=160\), or \(h=0.01\,\xi_{\rm hi}\) and \(h=0.003\,\xi_{\rm hi}\) in the activation
study and \(n_x=400\), \(h=0.01\,\xi_{\rm hi}\) in the strong-anisotropy cells; candidate spectra lie
on an equispaced grid over \([a,b]=[0.2,2.6]\) with step \(0.05\), or \(0.025\) in the cycle
experiments. Code, data and the ledger are archived at
\url{https://doi.org/10.5281/zenodo.23415148} (Section~P2 of the supplementary material).

The correspondence between each theoretical object and the evidence bearing on it, with the range
that evidence supports, is tabulated in Section~S11 of the supplementary material.

\subsection{Static spectral distortion and repair (EXP-1)}\label{sec:10.2}

\emph{Question.} Does resampling distort the spectrum as predicted, and does the plug-in inversion
recover the pre-distortion spectrum? The configuration uses
\(H_0=0.7\delta_{0.5}+0.3\delta_{2.5}\), \(c=0.75\), \(N\in\{500,1000,2000\}\), \(R=50\), with Gamma
weights for \(\alpha\in\{10,1,0.3\}\), all exogenous and satisfying (G4).

\emph{Forward law.} The post-resampling spectra match \(H_0\boxtimes\mathcal{T}_c(\nu)\) at every
level (Fig.~\ref{fig:2}(a)), including the zero-atom mass: \(0.33325\) against \(0.33333\) for
multinomial and \(0.24099\) against \(0.24311\) for residual. The second spectral moments follow
Corollary~\ref{cor:2} (Fig.~\ref{fig:2}(b)), four of six groups inside two standard errors with a
small negative bias for the two heavy-tailed groups.

\emph{Persistent bias and inversion.} The no-correction baseline converges to
\(W_1(H_0\boxtimes T,H_0)\), positive and not decreasing with \(N\) (Fig.~\ref{fig:2}(c)). All three
inverters reduce the error by an order of magnitude: at \(\alpha=1\) the bootstrap log-log slopes are
\(-0.49\) and \(-0.40\) (oracle), \(-0.43\) and \(-0.61\) (conditional plug-in), \(-0.65\) and
\(-1.02\) (realized count) for multinomial and residual resampling, all with confidence intervals
bounded away from zero, so the realized-count inverter decays faster than \(N^{-1/2}\). At \(N=2000\)
the errors are \(0.769/0.705\) without correction against \(0.034/0.020\) for the realized-count
inverter. Ledoit--Wolf nonlinear shrinkage \cite{ref15} improves on the uncorrected spectrum but stays
a factor of \(4.3\) to \(6.2\) behind every structure-aware inverter, because it assigns uniform mass
to the null space where the count-rank identity restores the spectrum exactly; the baseline is
model-free, so the gap measures the value of the count-law structure rather than a like-for-like
contest.

\begin{center}
\includegraphics[width=0.9\linewidth]{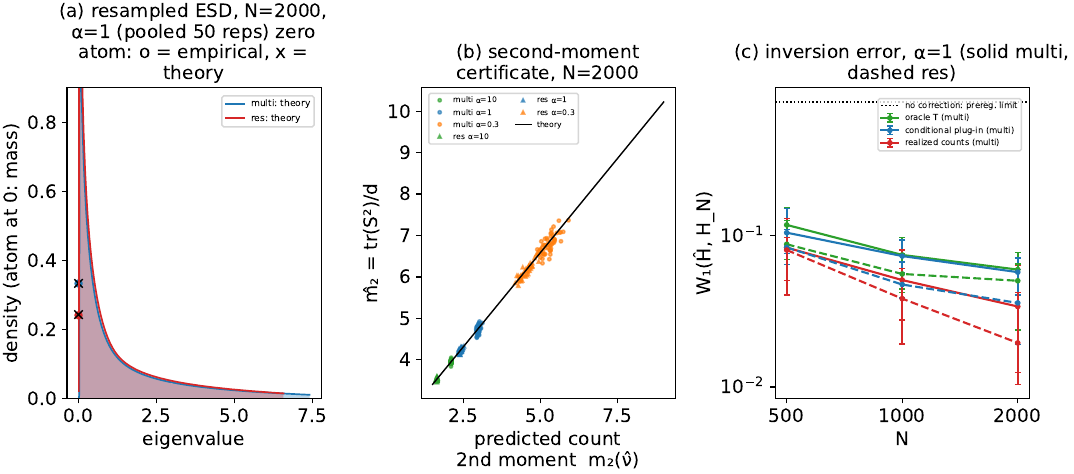}
\captionof{figure}{Static spectral distortion and repair (EXP-1). (a) Post-resampling ESD against $H_0\boxtimes\mathcal{T}_c(\nu)$ for both schemes at $N=2000$, $\alpha=1$; markers compare the empirical zero atom (o) with the closed form (x) [Theorem~\ref{thm:3}, Corollary~\ref{cor:1}]. (b) Second spectral moment against the predicted count moment for the six groups; the line is Corollary~\ref{cor:2}(4). (c) Median $W_1$ inversion error against $N$ for the three inverters (solid: multinomial; dashed: residual; bars: interquartile range); the dotted line is the limit of the uncorrected spectrum [Theorem~\ref{thm:4}].}\label{fig:2}
\end{center}

\subsection{Identifiability and conditioning diagnostics (EXP-2)}\label{sec:10.3}

\emph{Mismatch diagnostic.} Nine settings are compared at \(N=2000\), \(c=0.75\), \(R=50\) (Fig.~\ref{fig:3}(a)); \(p\)-values are one-sided Mann--Whitney tests against the correct multinomial baseline, whose median statistic is \(2.55\times10^{-3}\). The statistic stays at this level for correct residual resampling (\(2.27\times10^{-3}\), \(p=0.88\)) and fires in two settings: the wrong mechanism (\(2.76\times10^{-2}\), \(p=3.5\times10^{-18}\)) and strongly correlated particles at \(\rho=0.9\) (\(5.61\times10^{-3}\), \(p=4.4\times10^{-16}\)), which are the two settings where the median inversion error grows most, by factors of \(5.5\) and \(4.0\). It is dormant elsewhere. For endogenous weights depending on a \(q\)-dimensional projection with \(q=10\), covered by Proposition~S2 of the supplementary material, the error is unchanged and there is nothing to detect; at \(q=750\), where \(q/d=0.5\) lies outside that proposition, and for \(\rho=0.5\), dormancy coincides with a moderate error increase, by factors of \(1.22\) and \(1.35\). This is the scope of Theorem~\ref{thm:5}: the diagnostic tests consistency with the forward model, not estimation error, and a mild violation that keeps the observed spectrum close to the image of the forward map shifts the estimate without raising the residual (Section~\ref{sec:6.4}). It detected every violation that inflated the error severalfold and missed only those that inflated it by at most \(35\%\).

\begin{center}
\includegraphics[width=0.85\linewidth]{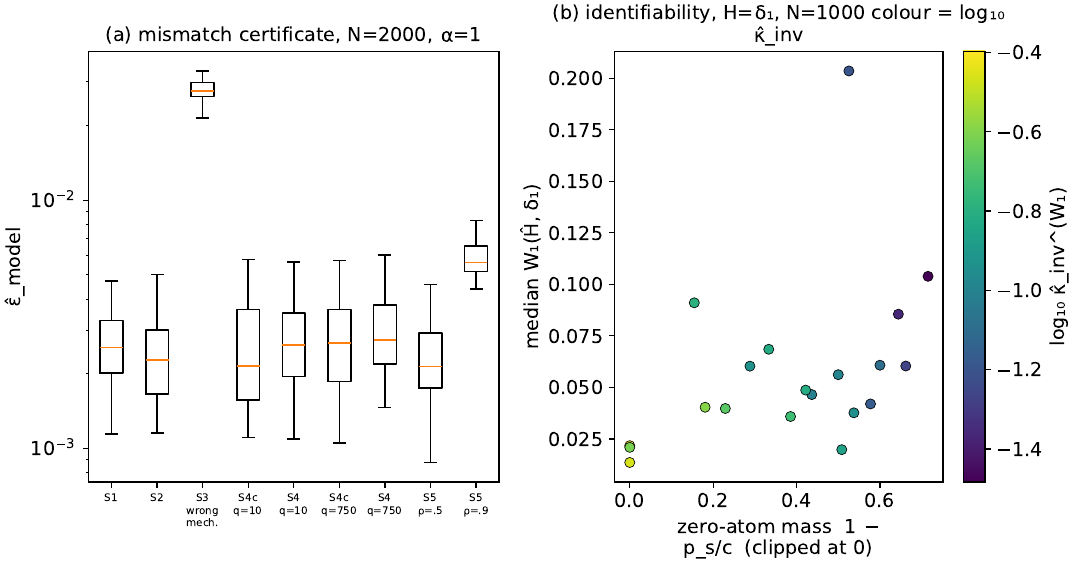}
\captionof{figure}{Diagnostics for EXP-2. (a) Mismatch statistic across the nine settings against the correct multinomial baseline [Theorem~\ref{thm:5}; a consistency diagnostic, not an exogeneity test]. (b) Inversion error against the zero-atom mass, colored by the estimated condition number [Theorem~\ref{thm:6}, Remark~\ref{rem:3}].}\label{fig:3}
\end{center}

\emph{Identifiability.} The condition number is driven by the zero-atom mass: the Spearman
correlation of \(\widehat\kappa_{\rm inv}\) is \(-0.931\) against that mass and \(-0.817\) against
\(|c-p_s|\), and that of the inversion error against \(\widehat\kappa_{\rm inv}\) is \(-0.684\). At the
boundary \(c\approx p_s\) the inversion is well conditioned, which confirms that the boundary is where
the atom is born rather than the mechanism of ill-conditioning (Remark~\ref{rem:3}); the pre-registered
prediction was revised accordingly.

\emph{Where conditioning enters.} At the map level the condition number degrades toward large \(c\) and
heavy-tailed weights, where the zero atom is largest; the discretization, optimizer and bandwidth of the
pipeline amplify that degradation. Section~S-K of the supplementary material separates the two levels on the two-point candidate
family in floating point, not as certified bounds, and no constant from it enters any theorem. The
comparative clause activates only on part of the \((c,\alpha)\) grid, not at the deployment setting of
EXP-3 (Section~\ref{sec:10.5}).

\subsection{Rejuvenation and stable full-cycle behavior (EXP-3)}\label{sec:10.4}

\begin{center}
\includegraphics[width=0.90\linewidth]{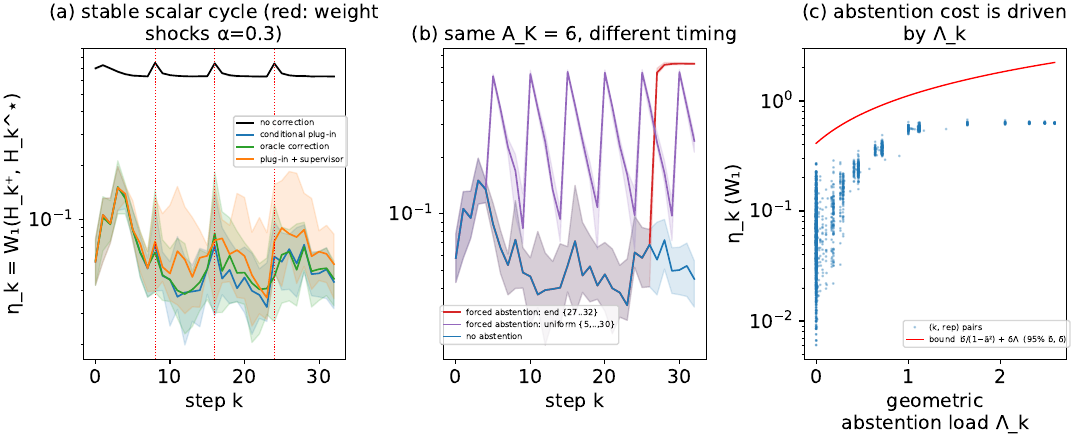}
\captionof{figure}{Full-cycle behavior (EXP-3). (a) Median error $\eta_k=W_1(H_k^+,H_k^{\star,+})$ over 20 repetitions for the four methods; red lines mark the weight shocks [Theorem~\ref{thm:8}(a)]. (b) Forced abstention at equal count $A_K=6$, clustered or spread. (c) Error against the geometric abstention load $\Lambda_k$ with the bound of Proposition~\ref{prop:3}.}\label{fig:5}
\end{center}

\emph{Question.} Does the corrected spectrum improve Metropolis proposal geometry without moving
the target, and does the full-cycle error accumulate? The cycle uses \(x_{k+1}=0.8x_k+\xi_k\),
\(\xi_k\sim\mathcal{N}(0,0.2I_d)\), \(d=750\), \(N=1000\), \(k=0,\dots,32\), \(R=20\), with
exogenous weight schedules \(\alpha_k=3\) and degeneration shocks at \(k\in\{8,16,24\}\).
Table~\ref{tab:1} reports the Metropolis preconditioning results.

The median \(\eta_k=W_1(H_k^+,H_k^{\star,+})\) over the 20 repetitions (Fig.~\ref{fig:5}(a)) is
\(0.698/0.749/0.629\) (initial/peak/final) without correction, against \(0.058/0.150/0.045\) for conditional plug-in, \(0.059/0.152/0.047\) for oracle and \(0.058/0.150/0.056\) for plug-in with supervisor, the last with 26 abstentions all from the mismatch clause. Corrected trajectories are bounded, recover within about three steps after each shock, and satisfy Theorem~\ref{thm:8}(a) on 100\% of steps when the per-step term is the realized correction error \(W_1(\widehat H_k,H_k^-)\) against the known population spectrum, which checks the propagation structure, not the estimated budget \(\widehat{\mathcal{B}}_N\) (Remark~\ref{rem:2}). The uncorrected trajectory stagnates at a positive plateau, the predicted compounding-versus-resetting contrast discussed after Proposition~\ref{prop:3}

Between shocks the corrected trajectories settle near \(0.045\) and peak near \(0.15\). The plateau
is governed by the fixed-\(N\) per-step budgets, whose median realized value is \(0.048\), not by the
resolution floor of Theorem~\ref{thm:8}(b$^{\prime}$); the resolution scale enters that theorem only
through an upper bound, so no value of \(\theta\) is inferred from these data.

\emph{Abstention-cost law.} The forced-abstention control uses identical counts \(A_K=6\) in two
schedules: clustered at the end (geometric load \(\Lambda_K=2.587\), median \(\eta_{32}=0.631\)) against
spread (\(\Lambda_K=0.459\), median \(0.244\)), significant at \(p=9.5\times10^{-7}\) (one-sided
Wilcoxon) with all 20 repetitions in the predicted direction. The cost of abstention is governed by the
geometric load, not by the count, which verifies Proposition~\ref{prop:3} directly.

\begin{center}

\captionof{table}{Metropolis preconditioning in Regime A ($d=225$, $c=0.15$, $\kappa(\Sigma)=50$, $r_{\rm obs}=200$): condition number of the preconditioned posterior covariance, slowest-direction ESS per step, acceptance rate, ESS per second, and maximum $|z|$ of the target-displacement test. Target preservation and mixing constants: Lemma~\ref{lem:6}, Theorem~\ref{thm:7}.}\label{tab:1}

\begin{tabularx}{\linewidth}{@{}Xrrrrr@{}}
\toprule
Preconditioner & $\kappa$ & ESS/step & Acc. & ESS/s & $\max|z|$\\
\midrule
$I$ & 42.4 & $3.26\times10^{-4}$ & 0.237 & 1355 & 3.23\\
$S_r^c+\varepsilon I$ & 19.6 & $5.45\times10^{-4}$ & 0.226 & 2242 & 1.60\\
lift$(\widehat H)+\varepsilon I$ & 5.6 & $8.72\times10^{-4}$ & 0.237 & 2046 & 1.88\\
oracle $\Sigma+\varepsilon I$ & 2.4 & $9.72\times10^{-4}$ & 0.240 & 1794 & 2.41\\

\bottomrule
\end{tabularx}
\end{center}

Regime A uses $d=225$, $c=0.15$, $\kappa(\Sigma)=50$, $r_{\rm obs}=200$. At matched acceptance ($\approx0.237$, close to the asymptotically optimal $0.234$ for random-walk proposals \cite{ref47}), the lifted preconditioner attains approximately 90\% of the oracle's slowest-direction ESS per step, so the gains appear in mixing efficiency rather than in acceptance; the global null of no target displacement is not rejected at the 0.05 level (\v{S}id\'ak-adjusted $p=0.059$).

\subsection{Supervisor mechanics and robustness}\label{sec:10.5}

The comparative clause is a safety net rather than an active rule: its activation region lies at
mid-to-high \(c\), above the deployment setting of EXP-3, so it never fires and has no true positives
there, and all 26 abstentions are mismatch detections (Section~\ref{sec:10.3}). An earlier standalone
condition-number threshold abstained 127 times, clustered at the shock steps; the comparative rule
reduced this to 26 without changing the peak error of \(0.150\).

Three measured qualifications apply. The monitor experiment is negative: under degeneration shocks
the ESS falls by a factor of \(3.23\) in synchrony with the zero-atom mass, contradicting the
pre-registered prediction of ESS-blind spectral alarms, so the spectral diagnostics complement the ESS
rather than replace it. The centred-statistic slope test of Corollary~\ref{cor:2}(4) fails at \(-8\%\)
against a \(2\%\) tolerance, because heavy-tailed weights amplify the rank-one centering gap of
Lemma~\ref{lem:1}(iv); the conditional-moment tests remain unbiased. Finally, a fixed schedule
\(\gamma_N=\gamma_0/N\) does not adapt to the scale of \(H\) and degraded the strong-anisotropy fit to
\(W_1=2.51\) against \(0.19\), which is why the scale-free schedule is recommended.

\subsection{The certificate as an online monitor}\label{sec:10.6}

The mismatch residual of Theorem~\ref{thm:5} is not a post hoc summary. It is recomputed at every step
from quantities the filter already holds, so it can be logged and watched while the filter runs, at no
additional asymptotic cost. Fig.~\ref{fig:monitor} shows the single supervised run of EXP-3 read this
way: the residual rises as the ensemble degenerates and falls at each weight shock,
where the mixture of information across the shock makes the observed spectrum locally easier to
represent. The residual tracks the estimated condition number of panel (c), which is the mechanism of
Remark~\ref{rem:3}, and it does so without access to the population spectrum: the drop at each shock is
computed from the weights and the observed spectrum alone.

Two limits of this reading are recorded. First, the residual is a consistency statistic, not an error
estimate: at every step of the run the median residual is between \(0.0047\) and \(0.0119\), while
the actual error \(\eta_k\) ranges over an order of magnitude more. A rising residual says that the
forward model is being strained, not how far the estimate has moved. Second, the residual is not an
alarm in the sense the negative result of Section~\ref{sec:10.5} rules out: it responds to weight
degeneration as the ESS does, so on these data the two are redundant as alarms and complementary only
in what they measure. What the step-by-step record adds is the \emph{object} of the alarm. The ESS
reports that the ensemble has degenerated; the residual reports that the distortion law the repair
divides out is no longer being matched, which is the quantity the repair itself depends on.

\begin{center}
\includegraphics[width=0.72\linewidth]{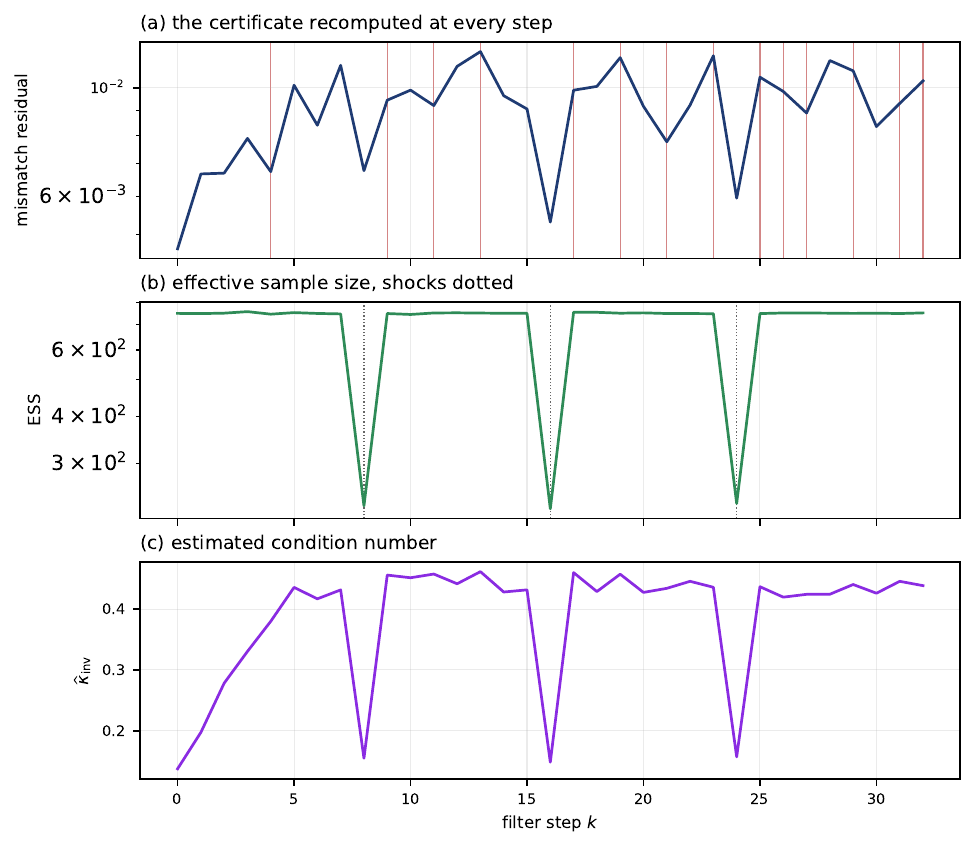}
\captionof{figure}{The mismatch certificate as an online quantity, from the supervised run of EXP-3.
(a) The mismatch residual recomputed at each filter step, with the steps at which the
supervisor abstained marked. (b) Effective sample size; dotted lines mark the weight shocks.
(c) Estimated condition number \(\widehat\kappa_{\rm inv}\). The residual and the condition number
move together, and both respond to the shocks, without either having access to the population spectrum.}
\label{fig:monitor}
\end{center}

\subsection{Departures from Gaussian fluctuations (EXP-4)}\label{sec:10.7}

Every theorem above assumes Gaussian entries in (G3), and universality is left open. This experiment
does not prove universality; it locates the boundary empirically. The design changes one object, the law
of the fluctuation matrix \(Z_N\), and holds the weights, the realized counts and the estimation grid
fixed, so the comparison is paired. Four laws are used: Gaussian, Rademacher (\(\pm1\)), and Student
\(t\) with five and three degrees of freedom, each rescaled to unit variance so that (G2) still
holds.

The forward law survives all four (Fig.~\ref{fig:disturbance}(a)). At \(N=2000\) the empirical
zero-atom mass lies within \(0.0052\) of the closed form for every law and both schemes, and in
particular does not drift as the tail is made heavier; the second spectral moment stays within
\(2.4\%\) of Corollary~\ref{cor:2}(4) throughout, against \(0.5\%\) for Gaussian entries. The
repair degrades in a way that is specific to the law (Fig.~\ref{fig:disturbance}(b)). Against the
Gaussian reference at the same seed, the median inversion error of the conditional plug-in inverter
under multinomial resampling is essentially unchanged for
Rademacher entries (ratio \(0.99\) at \(N=2000\)) and for \(t_5\) (\(1.01\)), but rises by a
factor of \(1.95\) for \(t_3\) at \(N=500\) and \(2.46\) at \(N=2000\). The factor does not
decrease over this range, so no rate is claimed: the degradation is a property of the finite-\(N\)
fluctuation law, and the consistency of Theorem~\ref{thm:4} concerns the pair
\((\varepsilon_N,\delta_N)\) under (G3) rather than this ratio.

The certificate does not report this degradation (Fig.~\ref{fig:disturbance}(c)). Its median is
within \(8\%\) of the Gaussian value for every law at both ensemble sizes, and the largest deviation,
for \(t_3\), is a small \emph{decrease}. This is the documented scope of Theorem~\ref{thm:5}
rather than a defect: the residual measures mismatch \emph{between} the observed spectrum and the image
of the forward map, and a heavy-tailed fluctuation law that leaves the limiting spectrum close to that
image is not a mismatch in this sense. The experiment therefore delimits the claim. The forward law is
robust to the law of \(Z_N\) at the level of the limiting spectrum; the quantitative budget is not,
since it inherits a factor of about two under \(t_3\); and the certificate is a test of the
distortion law, not of the fluctuation law, so it stays silent exactly where the second bullet of
Section~\ref{sec:11} says it must.

\begin{center}
\includegraphics[width=0.90\linewidth]{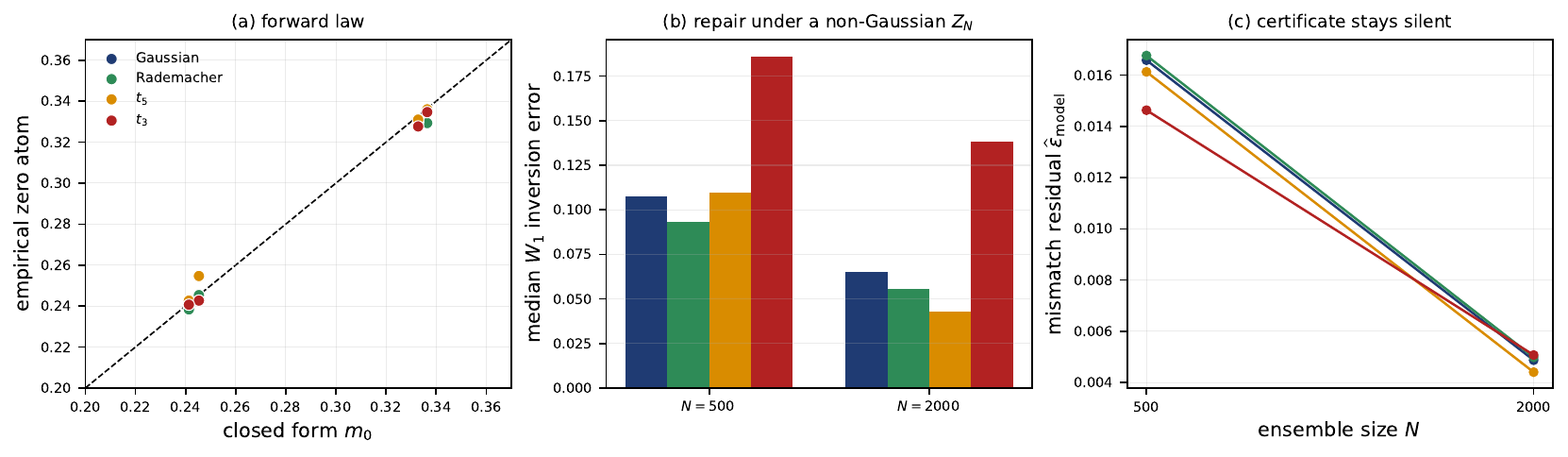}
\captionof{figure}{Departures from Gaussian fluctuations (EXP-4), paired by seed, weights and counts.(a) Empirical zero atom against the closed form of Corollary~\ref{cor:2}(1), all four laws and both schemes, at \(N=500\) and \(N=2000\); the dashed line is equality. (b) Median \(W_1\) inversion error per law at both ensemble sizes, conditional plug-in inverter, multinomial resampling. (c) The mismatch residual against ensemble size; the four curves agree to within \(8\%\), so a departure from Gaussianity is not what the certificate detects.}
\label{fig:disturbance}
\end{center}

\section{Limitations and conclusion}\label{sec:12}

\subsection{Boundaries of applicability}\label{sec:11}

\begin{enumerate}
\item \textbf{Gaussian fluctuations.} Theorem~\ref{thm:3} is proved for Gaussian entries; universality under finite $(4+\delta)$ moments is open, and Section~\ref{sec:10.7} locates the boundary empirically: the limiting spectral law survives Rademacher and Student-$t$ entries, the quantitative error budget does not, inheriting a factor of about two under $t_3$, and the mismatch certificate does not respond because it tests the distortion law rather than the fluctuation law.
\item \textbf{Exogeneity.} Assumption (G4) is indispensable in general: likelihood-driven weights coupled to the full state break the two-channel factorization. A partial relaxation is proved separately for weights depending on a low-dimensional projection, and the mismatch diagnostic is not a test of exogeneity: it can stay dormant where the inversion error rises moderately (Section~\ref{sec:10.3}).
\item \textbf{Finite resolution and dynamics.} Quantitative statements hold at the resolution scale $\theta$, with the certified full-cycle floor $\theta/(1-\bar a^2)$; the closed recursion is proved for scalar stable dynamics, time-uniform claims require a time-uniform envelope, and stability under general or unstable dynamics is not claimed: for $\bar a\ge1$ the error accumulates unless $\mathcal{B}_k$ decays at a compensating rate.
\item \textbf{Preconditioner regime.} The lifted preconditioner pays off only at low $c$ with nontrivial posterior anisotropy; outside that regime the identity preconditioner is competitive or better.
\end{enumerate}

Resampling repairs weight degeneracy at a spectral cost that is, in the high-dimensional regime, computable. This paper shows that the cost is also reparable: within an explicitly identifiable and exogenous regime, resampling-induced spectral distortion can be regularized and locally inverted, with an explicit local error budget and a certified mismatch diagnostic. The repaired spectrum is used only as a Metropolis-adjusted preconditioner, so it can improve proposal geometry without changing the invariant target. Under scalar stable dynamics the full-cycle error budget is geometrically summable, with the certified bound descending to the resolution floor at which the conditioning statement holds; under general dynamics full-cycle stability is not claimed, and we instead quantify the cost of supervisory abstention through the geometric abstention load. The corrected filtering step of Section~\ref{sec:9.1} and the pre-registered experiments of Section~\ref{sec:10} support these claims at every level at which they are stated: the forward distortion is computable, the repair is consistent and budgeted, the mismatch is certifiable, and the supervisor's abstentions are safe and costed.

\funding{This work was supported by the Science and Technology Planning Project of Guangdong Province, China [grant number 2025A0505010006]; the Joint Research Funding Program between the Macao Science and Technology Development Fund (FDCT) and the Department of Science and Technology of Guangdong Province [grant number 0108/2024/AGJ]; and Macao Polytechnic University [grant number RP/FCA-09/2025]. The funders had no role in study design, data collection and analysis, interpretation, writing of the report, or the decision to submit the article for publication.}

\section*{CRediT authorship contribution statement}
\textbf{Deming Song:} Conceptualization, Methodology, Software, Formal analysis, Investigation, Writing -- original draft, Visualization.
\textbf{Chi-Kin Lam:} Supervision, Funding acquisition, Writing -- review \& editing, Project administration.
\textbf{Kefeng Li:} Validation, Writing -- review \& editing.
\textbf{Hui Yu:} Resources, Validation.
\textbf{Xilin Li:} Resources, Validation.
\textbf{Huagang Lin:} Resources, Validation.

\data{Pre-registered code, data snapshots, and the deviation ledger are archived at Zenodo, \url{https://doi.org/10.5281/zenodo.23415148}. The implementation and experiment scripts are available at \url{https://github.com/songming-teaching/spectral-repair-resampling}.}

\suppdata{The supplementary material is organised in three parts. Part~I gives the proofs of every result stated in the main text, in Sections~S-A to S-L, and follows the order of the theorems: the centering lemma (S-A), the count laws (S-B), continuity of the spectral map (S-C), the random-environment spectral law with its two corollaries (S-D), the structure of the forward map and the plug-in lemma, including the Proposition~S1 modulus bound and the analysis of where the degradation lives (S-E), and then the consistency budget, the mismatch certificate and the target-preservation and full-cycle proofs (S-F to S-I). Sections~S-K and S-L carry the numerical conditioning study on the two-point family and the low-rank relaxation of the exogeneity assumption. Part~II holds the experimental sections S1 to S11, with the count-law checks, the complete comparison for Corollary~2, the inversion sensitivity and fallback studies, the Ledoit--Wolf baseline, the phase diagrams and supervisor audit table, the Metropolis--Hastings details, the activation zone of the comparative clause, the out-of-scope stress test with endogenous weights, the passive monitor experiment, and the theory--evidence correspondence table of Section~S11. Part~III contains the solver validation protocol (P1), the reproducibility protocol (P2) and the complete step-by-step specification of the filtering step (P3). Theorem, lemma, proposition and section numbers without a prefix refer to the main text. In this arXiv version the supplement follows the references, and both parts use the same reference list.}
\section*{Declaration of competing interest}
The authors declare that they have no known competing financial interests or personal relationships that could have appeared to influence the work reported in this paper.

\bibliographystyle{unsrtnat}
\bibliography{refs}

\clearpage
\pdfbookmark[0]{Supplementary material}{supplementary-material}
\begin{center}
{\LARGE\bfseries Supplementary Material for\\[6pt]
{\Large\bfseries \papertitle}\par}
\vspace{12pt}
{\large Deming Song, Chi-Kin Lam, Kefeng Li, Hui Yu, Xilin Li and Huagang Lin\par}
\end{center}
\bigskip

\setcounter{figure}{0}
\setcounter{table}{0}
\setcounter{equation}{0}
\renewcommand{\thefigure}{S\arabic{figure}}
\renewcommand{\thetable}{S\arabic{table}}
\renewcommand{\theequation}{S\arabic{equation}}
\renewcommand{\theHfigure}{SM.\arabic{figure}}
\renewcommand{\theHtable}{SM.\arabic{table}}
\renewcommand{\theHequation}{SM.\arabic{equation}}

This supplement has three parts. Part I contains the detailed proofs of the results stated in the main text; its sections S-A to S-L are the targets of the main text's pointers to proofs. Part II contains the experimental sections S1 to S11, with the registered protocols and the complete numerical detail. Part III contains the solver validation, reproducibility and algorithm specifications P1 to P3. Theorem, lemma, proposition and section numbers without a prefix refer to the main text. The reference list is shared with the main text and appears immediately before this supplement. No statement in this supplement extends the scope of the main-text claims.

\part*{Part I. Detailed proofs}\addcontentsline{toc}{part}{Part I. Detailed proofs}

\section*{S-A. Proof of Lemma 1 (centering)}\label{sm:S-A}\addcontentsline{toc}{section}{S-A. Proof of Lemma 1 (centering)}

Recall \(\hat m = \sum_i w_i x_i = Xw\), \(\hat m_r = \frac1N X n\), \(C_n = \operatorname{Cov}(n\mid X,w)\), \(B_n = \operatorname{Cov}(\hat m_r\mid X,w) = \frac{1}{N^2}XC_nX^\top\).

\subsection*{Part (i)}

By (A2), \(\mathbb{E}[\hat m_r\mid X,w] = \frac1N X\,\mathbb{E}[n\mid X,w] = Xw = \hat m\). Also \[\mathbb{E}[S_r^u\mid X,w] = \frac1N\sum_i \mathbb{E}[n_i\mid X,w]\,x_ix_i^\top = \sum_i w_i x_ix_i^\top = S_w^u.\] Therefore \[\mathbb{E}[S_r^c\mid X,w] = \mathbb{E}[S_r^u\mid X,w] - \mathbb{E}[\hat m_r\hat m_r^\top\mid X,w] = S_w^u - \hat m\hat m^\top - \operatorname{Cov}(\hat m_r\mid X,w) = S_w^c - B_n.\] \(B_n \succeq 0\) as a quadratic form in a covariance matrix, so \(\mathbb{E}[S_r^c\mid X,w] \preceq S_w^c\) in Loewner order.

\emph{Trace magnitude.} For multinomial resampling \(C_n = \operatorname{diag}(\lambda) - \frac1N\lambda\lambda^\top\), so \[\operatorname{tr}B_n = \frac1N\left[\frac1N\sum_i \lambda_i\|x_i\|^2 - \|\hat m\|^2\right].\] Under the stated condition \(\frac1N\sum_i\lambda_i\|x_i\|^2 = O_p(d)\), \(\operatorname{tr}B_n = O_p(d/N) = O_p(1)\) and \(d^{-1}\operatorname{tr}B_n = O_p(1/N)\). The matrix \(B_n\) may have rank up to \(d\); it is not low-rank. \(\square\)

\subsection*{Part (ii)}

Conditionally on \((X,w)\), \(B_n\) is deterministic and \(M := S_r^c + B_n\) satisfies \(\mathbb{E}[M\mid X,w] = S_w^c\) by rearranging part (i). Jensen's inequality applied to the matrix-valued random element \(M\) and convex \(\Phi\) gives \(\Phi(\mathbb{E}[M\mid X,w]) \le \mathbb{E}[\Phi(M)\mid X,w]\). Admissible functionals include \(\Phi(M) = \operatorname{tr}M^2\), \(\lambda_{\max}(M)\), \(-\log\det(M+\varepsilon I)\) (\(\varepsilon>0\)), and \(M \mapsto \operatorname{tr}\varphi(M)\) for scalar \(\varphi\) convex on the relevant interval with \(\operatorname{tr}\varphi\) finite. \(\square\)

\subsection*{Parts (iii) and (iv)}

\(S_r^u - S_r^c = \hat m_r\hat m_r^\top\) has rank at most one, and likewise on the weighted side. The rank inequality for empirical spectral distributions of Hermitian matrices, \(\sup_x|F^A(x) - F^B(x)| \le \operatorname{rank}(A-B)/d\), gives \(\sup_x|F^{S_r^c}(x) - F^{S_r^u}(x)| \le 1/d\). Hence (a) the limiting spectral distributions exist or fail to exist together and coincide when they exist; (b) the zero-atom masses differ by at most \(1/d\), since by interlacing a rank-one perturbation moves at most one zero eigenvalue. \(\square\)

\emph{Boundary note.} Finite-rank perturbations can create, delete, or move outliers (for example \(I_d\) against \(I_d + d\,e_1e_1^\top\)); the statements do not cover extreme eigenvalues or unbounded spectral moments. Edge convergence transfers only if one separately proves the absence of finite-rank outliers and a common limit of adjacent extreme order statistics (for example \(\lambda_1(S_r^u), \lambda_2(S_r^u) \to b_+\)), in which case interlacing \(\lambda_{j+1}(S^u) \le \lambda_j(S^c) \le \lambda_j(S^u)\) yields the same edge for the centered matrix. Neither \(\sup_N\|\Sigma_N\|_{\rm op} < \infty\) nor moment control alone implies this condition, as spiked models show.

\section*{S-B. Proof of Theorem 1 (count laws under random weights)}\label{sm:S-B}\addcontentsline{toc}{section}{S-B. Proof of Theorem 1 (count laws under random weights)}

\subsection*{Lemma S1 (balls into bins and Efron--Stein)}

Conditionally on \(w\), write multinomial counts as \(n_i = \sum_{\ell=1}^N \mathbf{1}\{Y_\ell = i\}\) with \(Y_\ell\) i.i.d. \(\mathrm{Categorical}(w)\). For bounded \(\phi\) set \(F(Y_1,\dots,Y_N) = \frac1N\sum_i \phi(n_i)\). Replacing one \(Y_\ell\) by an independent copy changes at most two coordinates of the count vector, hence \(|F - F^{(\ell)}| \le 4\|\phi\|_\infty/N\). The Efron--Stein inequality gives \[\operatorname{Var}(F\mid w) \le \frac12\sum_{\ell=1}^N \mathbb{E}\bigl[(F - F^{(\ell)})^2\mid w\bigr] \le \frac{8\|\phi\|_\infty^2}{N}.\] For residual resampling the same argument with \(R_N\) balls (categories \(u_i/R_N\)) gives \(\operatorname{Var} \le 8R_N\|\phi\|_\infty^2/N^2 = O_p(1/N)\), since \(R_N \le N\). No condition on \(\max_i\lambda_i\) and no pairwise covariance expansion is needed. \(\square\)

\subsection*{Part (i): multinomial resampling}

Write \(m(\phi;\tilde\rho) := \int \mathbb{E}[\phi(\mathrm{Poi}(\lambda))]\,\tilde\rho(d\lambda)\); the map \(\lambda \mapsto \mathbb{E}[\phi(\mathrm{Poi}(\lambda))]\) is bounded and continuous, since \(d_{\rm TV}(\mathrm{Poi}(\lambda),\mathrm{Poi}(\lambda')) \le |\lambda - \lambda'|\).

\emph{Expectation.} \(n_i\mid w \sim \mathrm{Bin}(N, \lambda_i/N)\), and Le Cam's bound \(d_{\rm TV}(\mathrm{Bin}(N,\lambda_i/N), \mathrm{Poi}(\lambda_i)) \le \lambda_i^2/N\) gives \[\left|\mathbb{E}\Bigl[\tfrac1N{\textstyle\sum_i}\phi(n_i)\,\big|\,w\Bigr] - m(\phi;\rho_N)\right| \le 2\|\phi\|_\infty \frac1{N^2}\sum_i\lambda_i^2 = O_p(1/N),\] and \(\rho_N \Rightarrow \rho\) in probability with a bounded continuous integrand gives \(m(\phi;\rho_N) \to \int\phi\,d\mathrm{MixPoi}(\rho)\) in probability.

\emph{Variance.} Lemma S1 gives \(\operatorname{Var} = O_p(1/N)\).

\emph{From conditional to unconditional convergence.} Take a countable convergence-determining class of bounded Lipschitz test functions; for each \(\phi\) apply the conditional Chebyshev inequality (the conditional bounds are deterministic functions of \(\rho_N\) vanishing in probability), and combine over the countable class by a finite \(\varepsilon\)-net and a diagonal argument. Tightness is immediate, since \(\frac1N\sum_i n_i = 1\) implies \(\nu_N([M,\infty)) \le 1/M\). The subsequence principle then yields \(d_{\rm BL}(\nu_N, \mathrm{MixPoi}(\rho)) \to 0\) in probability. \(\square\)

\subsection*{Part (ii): residual resampling}

The primitive assumption is \(\pi_N \Rightarrow \pi\) in probability. Since the \(u\)-coordinate projection is bounded and continuous, \(R_N/N = \frac1N\sum_i u_i \to \mathbb{E}_\pi U\) in probability.

\emph{Case 1: \(\mathbb{E}_\pi U > 0\).} Then \(R_N = \Theta_p(N)\) and \(\max_i u_i/R_N = O_p(1/N)\). Per coordinate, \(\tilde n_i\mid w \sim \mathrm{Bin}(R_N, u_i/R_N)\) with \(d_{\rm TV}(\cdot, \mathrm{Poi}(u_i)) \le u_i^2/R_N\), and the average over \(i\) is at most \(\frac1N\sum_i u_i/R_N = 1/N\). Hence for bounded continuous \(\phi\), \[\frac1N\sum_i \mathbb{E}[\phi(k_i + \tilde n_i)\mid w] = \frac1N\sum_i \mathbb{E}[\phi(k_i + \mathrm{Poi}(u_i))] + O(\|\phi\|_\infty/N) \to \mathbb{E}_\pi\,\phi(K + \mathrm{Poi}(U)),\] where the last step uses \(\pi_N \Rightarrow \pi\) and the bounded continuity of \((k,u) \mapsto \mathbb{E}\phi(k + \mathrm{Poi}(u))\). The variance vanishes by the residual version of Lemma S1.

\emph{Case 2: \(\mathbb{E}_\pi U = 0\).} Then \(R_N/N \to 0\). For bounded \(\phi\), \[\left|\int\phi\,d\nu_N - \frac1N\sum_i\phi(k_i)\right| \le 2\|\phi\|_\infty\,\frac1N\sum_i\mathbf{1}\{\tilde n_i \ge 1\},\] and \(\mathbb{E}[\frac1N\sum_i\mathbf{1}\{\tilde n_i\ge 1\}\mid w] \le R_N/N \to 0\) (using \(\mathbb{P}(\tilde n_i \ge 1\mid w) = 1 - (1 - u_i/R_N)^{R_N} \le u_i\)), so Markov's inequality concludes; meanwhile \(\frac1N\sum_i\delta_{k_i} \Rightarrow \mathcal{L}(K)\), consistent with \(\mathcal{L}(K + \mathrm{Poi}(U))\) since \(U = 0\) almost surely. \(\square\)

\subsection*{Part (iii): second count moments}

\emph{Joint factorial moments.} For \(i \neq j\), \(\mathbb{E}[(n_i)_a(n_j)_b\mid w] = (N)_{a+b}w_i^aw_j^b\), and \(n^2 = (n)_2 + (n)_1\).

\emph{Multinomial, expectation.} \(\mathbb{E}[n_i^2\mid w] = \lambda_i + \lambda_i^2(1 - 1/N)\), so \(\mathbb{E}[\frac1N\sum_i n_i^2\mid w] = 1 + \frac1N\sum_i\lambda_i^2 - \frac1{N^2}\sum_i\lambda_i^2 \to 1 + r_2\) in probability.

\emph{Multinomial, variance.} \(\operatorname{Var}(n_i^2\mid w) \le C(\lambda_i + \lambda_i^2 + \lambda_i^3)\) (compare the Poisson case, \(\operatorname{Var}(M^2) = 4\lambda^3 + 6\lambda^2 + \lambda\)). The diagonal sum is \((1/N^2)\sum_i\operatorname{Var}(n_i^2) = O_p(N^{-1/2})\), using \(\sum\lambda_i = N\), \(\sum\lambda_i^2 = O_p(N)\), and \(\sum\lambda_i^3 \le \max_i\lambda_i \cdot \sum\lambda_i^2 = O_p(N^{3/2})\). The joint factorial moments show that the leading terms of the off-diagonal covariances cancel: \(|\operatorname{Cov}(n_i^2,n_j^2)| = O\bigl((\lambda_i^2\lambda_j^2 + \lambda_i^2\lambda_j + \lambda_i\lambda_j)/N\bigr)\), which sums to \(O_p(1/N)\) at scale \(1/N^2\). Hence \(\operatorname{Var}(\frac1N\sum_i n_i^2\mid w) = O_p(N^{-1/2})\).

\emph{Residual.} With \(k_i = \lfloor\lambda_i\rfloor\), and noting that \(2k_iu_i\) is already inside \(\lambda_i^2 = k_i^2 + 2k_iu_i + u_i^2\), \[n_i^2 - \lambda_i^2 = 2k_i(\tilde n_i - u_i) + (\tilde n_i^2 - u_i^2).\] For the expectation, \(\mathbb{E}[\tilde n_i^2 - u_i^2\mid w] = u_i - u_i^2/R_N\) (from \(\mathbb{E}[\tilde n_i^2\mid w] = u_i + u_i^2(1 - 1/R_N)\)), so \[\mathbb{E}\Bigl[\tfrac1N{\textstyle\sum_i}n_i^2\,\big|\,w\Bigr] = \frac1N\sum_i\lambda_i^2 + \frac{R_N}{N} - \frac{1}{NR_N}\sum_i u_i^2,\] where the last term lies in \([0, 1/N]\) and the second tends to \(\mathbb{E}_\pi U\) in probability (to \(0 = \mathbb{E}_\pi U\) in Case 2). For the cross term, with \(a_i = 2k_i\), the exact linear-form variance for \(\tilde n \sim \mathrm{Multinomial}(R_N; u/R_N)\) is \(\operatorname{Var}(\sum a_i\tilde n_i\mid w) = \sum u_i a_i^2 - (\sum u_i a_i)^2/R_N \le \sum u_i a_i^2 \le 4\sum\lambda_i^2\), so the centered cross term is \(O_p(1/N)\) at scale \(1/N^2\) in both cases. For the quadratic term, \(Q_N := \sum_i \tilde n_i^2 = \sum_i (\tilde n_i)_2 + R_N\); given \(w\), \(R_N\) is deterministic, so the randomness sits in \(\sum_i(\tilde n_i)_2\). The joint factorial moments \(\mathbb{E}[(\tilde n_i)_a(\tilde n_j)_b\mid w] = (R_N)_{a+b}(u_i/R_N)^a(u_j/R_N)^b\) and \(u_i \le 1\) give \(\operatorname{Var}(Q_N\mid w) = O(R_N)\): the diagonal terms \(\operatorname{Var}((\tilde n_i)_2) \le \mathbb{E}\tilde n_i^4 = O(u_i)\) sum to \(O(R_N)\), and the covariances are \(O(u_i^2u_j^2/R_N)\), summing to at most \((\sum u_i^2)^2/R_N = O(R_N)\). Hence in Case 1, \(\operatorname{Var}(Q_N/N\mid w) = O_p(R_N/N^2) = O_p(1/N)\); in Case 2, \(Q_N/N \ge 0\) with conditional mean at most \(2R_N/N \to 0\), and Markov's inequality concludes with no variance bound needed. Assembling, \(\frac1N\sum_i n_i^2 = \frac1N\sum_i\lambda_i^2 + R_N/N + o_p(1) \to r_2 + \mathbb{E}_\pi U\) in both cases. \(\square\)

\section*{S-C. Proof of Lemma 2 (continuity of the spectral map)}\label{sm:S-C}\addcontentsline{toc}{section}{S-C. Proof of Lemma 2 (continuity of the spectral map)}

\emph{Statement.} If \(\eta_L \Rightarrow \eta\) with \(\int t\,\eta_L(dt)\) uniformly bounded, then \(\mathcal{T}_c(\eta_L) \Rightarrow \mathcal{T}_c(\eta)\).

\emph{Proof.} The companion fixed-point equation \(z = -\frac{1}{m(z)} + \int \frac{t}{1+ctm(z)}\eta(dt)\), on the Stieltjes branch of a probability measure, is stable in \(\eta\): for \(\mathrm{Im}\,z \ge v_0\) the integrand \(t \mapsto t(1+ctm)^{-1}\) is uniformly bounded and continuous in \((t, m)\), so the right-hand side is weakly continuous in \(\eta\); the bound \(|m_L(z)| \le 1/\mathrm{Im}\,z\) gives equicontinuity; a subsequence argument yields a limit point that solves the equation for \(\eta\); and uniqueness of the fixed point on the stated branch forces convergence of the whole sequence. Pointwise convergence of Stieltjes transforms together with tightness, which follows from \(\int x\,\mathcal{T}_c(\eta_L)(dx) = \int t\,\eta_L(dt)\) by the moment multiplicativity of the representation \(D_c(\eta \boxtimes \mathrm{MP}(1/c))\), gives weak convergence of the measures. \(\square\)

\section*{S-D. Proofs of Theorem 3 and Corollaries 1 and 2}\label{sm:S-D}\addcontentsline{toc}{section}{S-D. Proofs of Theorem 3 and Corollaries 1 and 2}

\subsection*{Theorem 2 (imported deterministic core)}

For a deterministic diagonal \(\widetilde D_N\) with entries in \(\mathbb{Z}_{\ge 0}\), ESD \(\Rightarrow \widetilde\eta\) (a probability law with mean at most \(1\)), \(\frac1N\sum_i \widetilde d_i^{\,2} \le K < \infty\), Gaussian \(Z_N\) independent of \(\widetilde D_N\), and \(\Sigma_N\) satisfying (G2), the ESD of \(\frac1N\Sigma_N^{1/2}Z_N\widetilde D_N Z_N^\top\Sigma_N^{1/2}\) converges in probability to \(H \boxtimes \mathcal{T}_c(\widetilde\eta)\). Gaussianity is used at exactly one point: conditionally on \(\widetilde D_N\), \(Q_N = \frac1N Z\widetilde D Z^\top\) is orthogonally invariant and hence asymptotically free from \(\Sigma_N\). Unbounded diagonal entries are handled by truncation and rank approximation before passing to the limit, with continuity of the fixed point. This is \cite[Theorem 12]{ref01}, proved in \cite[SM \S{}S7]{ref01}.

\subsection*{Theorem 3}

\emph{Step 1 (good event).} For \(\varepsilon > 0\) and constants \(K_1, K_2\), define the environment event \[\mathcal{E}_N^{\rm good} = \{d_{\rm BL}(\rho_N, \rho) < \varepsilon\} \cap \{\tfrac1N{\textstyle\sum_i}\lambda_i^2 \le K_1\} \cap \{d_{\rm BL}(\nu_N, \nu) < \varepsilon\} \cap \{\tfrac1N{\textstyle\sum_i}n_i^2 \le K_2\}.\] By (W2) and Theorem 1, \(K_1\) and \(K_2\) can be chosen so that \(\mathbb{P}(\mathcal{E}_N^{\rm good}) \to 1\) as \(N\to\infty\) and then \(\varepsilon \downarrow 0\).

\emph{Step 2 (truncation and rank control).} On \(\mathcal{E}_N^{\rm good}\) let \(D_N^{(L)} = \operatorname{diag}(\min(n_i, L))\). Then \[\lim_{L\to\infty}\limsup_{N\to\infty}\mathbb{P}\Bigl(\tfrac1N\#\{i : n_i > L\} > \varepsilon\Bigr) = 0,\] since \(\frac1N\#\{i : n_i > L\} \le L^{-2}\frac1N\sum_i n_i^2 \le K_2 L^{-2}\) on the good event. With \(\operatorname{rank}(S_r^u - S_r^{u,(L)}) \le \#\{i : n_i > L\}\), the ESD rank inequality gives \[\sup_x\left|F^{S_r^u}(x) - F^{S_r^{u,(L)}}(x)\right| \le \frac{\#\{i : n_i > L\}}{d} = \frac{N}{d}\cdot\frac1N\#\{i : n_i > L\} = O_p(L^{-2})\] in the double-limit sense (first \(N\to\infty\), then \(L\to\infty\)). The truncated count law \(\nu^{(L)} = \mathcal{L}(\min(J,L))\), \(J \sim \nu\), has mean at most \(1\) and uniformly bounded second moment.

\emph{Step 3 (Theorem 2 and continuity).} Conditionally on the extended environment \(\mathcal{E}_N = \sigma(\mathcal{W}_N, n_N)\), the diagonal \(D_N^{(L)}\) is deterministic and satisfies the premises of Theorem 2 on the good event. Conditioning on \(\mathcal{W}_N\) alone would not fix \(D_N\); the auxiliary randomness \(U_N\) with \((\mathcal{W}_N, U_N) \perp Z_N\) closes this gap (Remark 1). Hence for each fixed \(L\), \(\mu_{S_r^{u,(L)}} \Rightarrow H \boxtimes \mathcal{T}_c(\nu_N^{(L)})\) in probability as \(N\to\infty\). Lemma 2 is used twice: (i) \(\nu_N^{(L)} \Rightarrow \nu^{(L)}\) on the good event, by continuity of truncation; (ii) \(\mathcal{T}_c(\nu^{(L)}) \Rightarrow \mathcal{T}_c(\nu)\) as \(L \to \infty\), the means being uniformly bounded. Combining the three contributions, \[d_{\rm BL}\bigl(\mu_{S_r^u}, H \boxtimes \mathcal{T}_c(\nu)\bigr) \le O(L^{-2}) + (\text{$N$-limit term}) + (\text{$L$-limit term}) \to 0.\]

\emph{Step 4 (pullback to the random environment).} All convergences above are conditional on the environment, on the good event. By the subsequence characterization of convergence in probability, every subsequence has a further subsequence along which the environment converges almost surely ((W2) and Theorem 1 hold in probability and therefore almost surely along subsequences), and on it the conditional weak convergence holds almost surely; weak convergence of measures is metrized by \(d_{\rm BL}\), so \(\mu_{S_r^u} \Rightarrow H \boxtimes \mathcal{T}_c(\nu)\) in probability. \(\square\)

\subsection*{Corollary 1}

\(\operatorname{rank}(S_r^u - S_r^c) \le 1\), so \(\sup_x |F^{S_r^u} - F^{S_r^c}| \le 1/d \to 0\) by Lemma 1(iii), which composed with Theorem 3 gives the claim. The zero atom is inherited up to \(1/d\); extreme eigenvalues are not inherited (Lemma 1(iv)). \(\square\)

\subsection*{Corollary 2}

(1) The maximum-atom rule for free multiplicative convolution of probability measures on \(\mathbb{R}_{\ge0}\) gives \((H \boxtimes \mathcal{T}_c(\nu))(\{0\}) = \max\{H(\{0\}), \mathcal{T}_c(\nu)(\{0\})\}\), with the multinomial specialization \(\mathcal{T}_c(\nu)(\{0\}) = \max(0, 1 - (1-\nu(\{0\}))/c)\). (2) \(p_s := 1 - \nu(\{0\})\); under multinomial resampling \(\nu(\{0\}) = \mathbb{E}[e^{-\Lambda}] = \mathcal{L}_\rho(1)\), and the rank threshold is \(p_s \gtrless c\), the identity-population transition of \cite{ref01}. (3) The benchmark \((\sqrt{c} - \sqrt{p_s})^2\) is the Marchenko--Pastur benchmark with unit active columns, by positive-semidefinite monotonicity with \(D_N\) dominating the active-column projection and \(H = \delta_1\); no uniform positive lower bound exists for general \(\Sigma_N\) with \(\lambda_{\min}(\Sigma_N) \to 0\). (4) \(m_2(H \boxtimes \mathcal{T}_c(\nu)) = m_{2,H} + ca_1^2 m_2(\nu)\) for \(m_1(\nu) = 1\), the spectral-moment identity of \cite{ref01}, with \(m_2(\nu)\) supplied by Theorem 1(iii). \(\square\)

\section*{S-E. Proofs for Chapter 3: structure of the forward map and the plug-in lemma}\label{sm:S-E}\addcontentsline{toc}{section}{S-E. Proofs for Chapter 3: structure of the forward map and the plug-in lemma}

\subsection*{Lemma 3}

\emph{(i) Compactness.} \(\mathcal{H}_{a,b} = \mathcal{P}([a,b])\): a uniformly bounded support implies uniform integrability and tightness, and weak limits stay in \([a,b]\). Hence \(\mathcal{H}_{a,b}\) is weakly compact. \(\square\)

\emph{(ii) Continuity.} Weak continuity of \(\boxtimes\) on \(\mathcal{P}(\mathbb{R}_{\ge0})\), with the second argument fixed and of finite first moment, is standard \cite{ref27}; alternatively, apply the fixed-point stability of Lemma 2. The S-transform is not used here. \(\square\)

\emph{(iii) Injectivity.} For \(G \in \mathcal{H}_{a,b}\), \(m_1(G) \ge a > 0\), and \(m_1(T) = 1\) gives \(S_T(0) = 1/m_1(T) = 1 \neq 0\). Since \(S_{G \boxtimes T} = S_G S_T\) on a common domain, division recovers \(S_G\); since \(G\) is compactly supported, it is moment-determinate, and equality of S-transforms near the origin yields equality of all moments, hence of the measures. For unbounded-support \(T\) the S-transform is not analytic on a full neighborhood of the origin but is defined on a left interval \((-\delta, 0]\), and the identity theorem along the interval gives the same conclusion. \(\square\)

\subsection*{Proposition S1 (zero-free region and modulus bound for the companion transform)}

Let \(T \in \mathcal{P}(\mathbb{R}_{\ge0})\) with \(m_1(T) = 1\), with no bounded-support assumption, and let \(s_T(z) = \int \frac{T(dx)}{1 - zx}\) be its companion transform. The count laws arising from resampling (mixed Poisson laws, residual limits, and plug-in Poisson mixtures) have unbounded support, so the relevant region is the left half-disk rather than a disk around the origin. On \(\{ \operatorname{Re} z \le 0,\ |z| \le r \}\):

(i) (Analyticity and zero-free left region.) \(s_T\) is analytic on \(\mathbb{C} \setminus [0,\infty)\), since its poles \(1/x\), \(x \in \operatorname{supp}T\), lie on the nonnegative axis, and for \(\operatorname{Re} z \le 0\), \[\operatorname{Re} s_T(z) = \int \frac{1 - \operatorname{Re}(z)\,x}{|1 - zx|^2}\,T(dx) > 0,\] because the numerator satisfies \(1 - \operatorname{Re}(z)x \ge 1\) for all \(x \ge 0\). Hence \(s_T\) has no zeros on the closed left half-plane.

(ii) (Support-free modulus bound.) For \(\operatorname{Re} z \le 0\) and \(|z| \le r\), \[|s_T(z)| \ge \operatorname{Re} s_T(z) \ge \int (1 + rx)^{-2}\,T(dx) \ge \frac{1}{(1 + r\,m_1(T))^2} = \frac{1}{(1+r)^2},\] where the second inequality uses \(|1 - zx|^2 \le (1 + |z|x)^2 \le (1 + rx)^2\), and the third is Jensen's inequality for the convex function \(x \mapsto (1+rx)^{-2}\). For a Poisson-type law at \(r = 1\) the measured minimum of \(\operatorname{Re} s_T\) is \(0.596\), against the bound \(0.25\). \(\square\)

\emph{Remark (two-level bound).} The intermediate term \(\int(1+rx)^{-2}T(dx)\) is markedly tighter than its Jensen closure, because the Jensen step discards all higher-moment information. The closed form \((1+r)^{-2}\) is the portable bound, requiring only \(m_1(T)=1\), while the integral, available at the cost of one quadrature, is the operative bound whenever \(T\) is explicitly computable, as is the case for \(T = \mathcal{T}_c(\nu)\) in the pipeline.

\subsection*{Why no explicit inverse modulus is claimed}

The division step of the inversion takes place on the S-transform side, \(S_G = S_\mu / S_T\). For \(w\) in a left neighborhood of \(0\), with \(z = \chi_T(w)\), \(\chi_T = \psi_T^{-1}\) and \(\psi_T = s_T - 1\), \[|S_T(w)| = \left| \frac{z\,s_T(z)}{s_T(z) - 1} \right|_{z = \chi_T(w)},\] and Proposition S1(ii) controls the numerator, since \(\chi_T(w)\) lies in the left region for such \(w\). A quantitative lower bound on \(\kappa_{\rm inv}^{(\theta)}\) along this route is nevertheless unavailable, for two structural reasons. First, the moment-determinacy step runs in the forward direction only: on \(\mathcal{P}([a,b])\), \(|m_k(G_1) - m_k(G_2)| \le k b^{k-1} W_1(G_1, G_2)\), hence \(\sup |s_{G_1} - s_{G_2}| \le [r/(1 - br)^2]\,W_1(G_1, G_2)\) on the left region. This says that the forward map contracts transform differences, which is the source of ill-conditioning; inversion requires the reverse inequality, transform differences controlling \(W_1\) from below, and this fails in general even on compact support. Restricting to pairs with \(W_1 \ge \theta\) removes the diagonal but does not bound the transform difference from below. Second, the inverse-function factor enters inversely: the Lipschitz modulus of \(\chi_T = \psi_T^{-1}\) requires a lower bound on \(|\psi_T'|\), and although \(\psi_T'(0) = 1\) under \(m_1(T) = 1\), a region-wide lower bound is again a form of inverse quantitative continuity that the forward estimates do not imply. What holds rigorously is Theorem 6: \(\mathcal{F}_T\) is a homeomorphism of the compact class \(\mathcal{U}\) onto its image, its inverse is uniformly continuous, and for every \(\theta > 0\) there exists \(\delta_\theta > 0\) with \(\kappa_{\rm inv}^{(\theta)} \ge \delta_\theta/(b-a) > 0\).

A quantitative modulus would require (a) an embedding between the band of \(d_{\rm St}\) and the left region, (b) a resolution-aware inverse moment modulus, and (c) control of the inverse-function distortion through a lower bound on \(|\psi_T'|\). For the two-point distortion law \(T = q\delta_0 + (1-q)\delta_{\ell_q}\), \(\ell_q := (1-q)^{-1}\), one has \(\psi_T'(z) = (1 - \ell_q z)^{-2}\), so \(\inf|\psi_T'|\) over \(\{\operatorname{Re} z \le 0,\ |z| \le 0.5\}\) equals \((1 + \ell_q/2)^{-2}\), which decreases from \(0.444\) at \(q = 0\) to \(0.141\) at \(q = 0.7\). No constant from this program enters any theorem.

\subsection*{Where the degradation lives}

Three levels can be distinguished. (i) \emph{Definition level.} The forward map \(G \mapsto G \boxtimes T\) contracts the fixed-bandwidth Stieltjes discrepancy more strongly relative to \(W_1\) as the zero-atom mass grows; this is the metric pair that defines \(\kappa_{\rm inv}^{(\theta)}\), and the effect is the one measured in Section 5.3 of the main text. (ii) \emph{Chain level.} Propagating a data error through the inversion chain \(\psi_\mu \to \chi_\mu \to S_\mu \to S_G = S_\mu/S_T \to \chi_G \to \psi_G\) meets the amplification factors \(1/|\psi_\mu'|\) on the data side, \(1/|S_T|\) at the division, and \(1/|\chi_G'|\) on the side of \(G\). Only the first depends on \(T\). At the origin it does not: \(\psi_\mu'(0) = m_1(\mu) = m_1(G)\,m_1(T) = m_1(G)\) for every \(T\) with \(m_1(T) = 1\), so any dependence on the zero-atom mass enters through the behavior of \(\psi_\mu'\) away from the origin, over the left region on which the inverse \(\chi_\mu\) is taken. No quantitative value of this modulus is used or claimed. (iii) \emph{Benign division.} For the two-point distortion law, \(S_T(w) = 1 + (1-\ell_q)\chi_T(w)\), hence \(|S_T|^2 = (1+(\ell_q-1)a_\chi)^2 + (\ell_q-1)^2 b_\chi^2 \ge 1\) for \(\chi_T(w) = -a_\chi + ib_\chi\) in the left region, at every \(q\). The division therefore never amplifies, which is why an estimate assembled from S-side moduli alone cannot capture the degradation. The quantity \(\inf|\psi_T'|\) enters only the exact evaluation of \(S_T\); it is a computable proxy correlated with the zero-atom mass, not an error-propagation factor. Note that \(s_\mu\) is the companion transform, which is not multiplicative under \(\boxtimes\); the multiplicative law belongs to the S-transform.

\subsection*{Lemma 4 (plug-in lemma)}

(i) Multinomial: the argument is the expectation step of Theorem 1(i), but deterministic: \(\widehat\nu_N^{\rm multi} = \frac1N\sum_i \mathrm{Poi}(\lambda_i)\) has \(m(\phi; \widehat\nu_N^{\rm multi}) = \frac1N\sum_i \mathbb{E}\phi(\mathrm{Poi}(\lambda_i)) = m(\phi;\rho_N) \to m(\phi;\rho)\); its first moment is exactly \(1\), and its second moment is \(1 + \frac1N\sum_i\lambda_i^2 = O_p(1)\) by (W2). (ii) Residual: \(\widehat\nu_N^{\rm res}\) integrates the bounded continuous kernel \((k,u) \mapsto \mathbb{E}\phi(k + \mathrm{Poi}(u))\) against \(\pi_N \Rightarrow \pi\); its first moment is exactly \(1\) and its second moment is \(\frac1N\sum_i\lambda_i^2 + R_N/N = O_p(1)\). (iii) Lemma 2, whose hypotheses are the first-moment preservation and second-moment boundedness just verified, gives \(\widehat T_N \Rightarrow T\). (iv) Joint continuity of \((G,Q) \mapsto G \boxtimes Q\) on the relevant compact sets gives uniform continuity on \(\mathcal{H} \times (\{T\} \cup \{\widehat T_N\})\), hence \(\delta_N = \sup_{G\in\mathcal{H}} d_{\rm St}(G \boxtimes \widehat T_N, G \boxtimes T) \to 0\) in probability. \(\square\)

\section*{S-F. Proof of Theorem 4 (consistency and local error budget)}\label{sm:S-F}\addcontentsline{toc}{section}{S-F. Proof of Theorem 4 (consistency and local error budget)}

The assumptions are \(H_0 \in \mathcal{H}_{a,b}\), \(\mu = \mathcal{F}_T(H_0)\), \(R(H_0) < \infty\), and \(\varepsilon_N \to 0\), \(\delta_N \to 0\), \(\gamma_N \to 0\) in probability. An argmin exists, since the criterion is lower semicontinuous on the compact set \(\mathcal{H}\).

\emph{Step 1 (minimality gives the forward fit).} By the argmin property and \(d_{\rm St}(\widehat\mu_N, \mathcal{F}_{\widehat T_N}(H_0)) \le \varepsilon_N + \delta_N\), \[d_{\rm St}\bigl(\widehat\mu_N, \mathcal{F}_{\widehat T_N}(\widehat H_N)\bigr)^2 + \gamma_N R(\widehat H_N) \le (\varepsilon_N + \delta_N)^2 + \gamma_N R(H_0).\]

\emph{Step 2 (forward residual).} By the triangle inequality, \[d_{\rm St}\bigl(\mathcal{F}_T(\widehat H_N), \mathcal{F}_T(H_0)\bigr) \le \delta_N + \sqrt{(\varepsilon_N + \delta_N)^2 + \gamma_N R(H_0)} + \varepsilon_N \to 0 \quad \text{in probability.}\]

\emph{Step 3 (consistency by the homeomorphism argument).} \(\mathcal{H}\) is compact and \(\mathcal{F}_T\) is continuous and injective, so \(\mathcal{F}_T : \mathcal{H} \to \mathcal{F}_T(\mathcal{H})\) is a continuous bijection from a compact space onto a Hausdorff space, hence a homeomorphism, and its inverse is uniformly continuous on the compact image. Therefore the convergence of Step 2 implies \(W_1(\widehat H_N, H_0) \to 0\) in probability; continuous maps preserve convergence in probability, and no subsequence extraction is needed. \(\square\)

\emph{Step 4 (the budget; no circularity).} Neighborhood entry: \(\Delta_{\mathcal{U}} := \inf_{G \in \mathcal{H}\setminus\mathcal{U}} d_{\rm St}(\mathcal{F}_T(G), \mathcal{F}_T(H_0)) > 0\), since \(\mathcal{H}\setminus\mathcal{U}\) is compact in the relative topology, \(\mathcal{F}_T\) is continuous, and injectivity makes the function pointwise positive; a vanishing forward residual therefore forces \(\mathbb{P}(\widehat H_N \in \mathcal{U}) \to 1\). On that event, distinguish \(W_1(\widehat H_N, H_0) \ge \theta\) and \(W_1(\widehat H_N, H_0) < \theta\). In the first case the definition of \(\kappa := \kappa_{\rm inv}^{(\theta)}(H_0,T;\mathcal{U})\) applies to the pair \((\widehat H_N, H_0)\) and gives \[W_1(\widehat H_N, H_0) \le \kappa^{-1}\left[\delta_N + \sqrt{(\varepsilon_N + \delta_N)^2 + \gamma_N R(H_0)} + \varepsilon_N\right] =: \mathcal{B}_N;\] in the second case \(W_1(\widehat H_N, H_0) < \theta\) trivially. Hence \(W_1(\widehat H_N, H_0) \le \max\{\theta, \mathcal{B}_N\}\). The condition number is evaluated at the true pair \((H_0, T)\), while the estimate is computed at the realized pair, so no circularity arises. Moreover \(\mathcal{B}_N \le \kappa^{-1}[2\varepsilon_N + 2\delta_N + \sqrt{\gamma_N R(H_0)}]\) by \(\sqrt{x^2+y^2} \le x+y\), and the rate statement follows by substitution. \(\square\)

\section*{S-G. Proof of Theorem 5 (consistent mismatch certificate)}\label{sm:S-G}\addcontentsline{toc}{section}{S-G. Proof of Theorem 5 (consistent mismatch certificate)}

With \(\varepsilon_{\rm model} := \inf_{G\in\mathcal{H}} d_{\rm St}(\mu, G \boxtimes T)\) and \(\widehat\varepsilon_{{\rm model},N} := \inf_{G\in\mathcal{H}} d_{\rm St}(\widehat\mu_N, G \boxtimes \widehat T_N)\), the inequality \(|\inf f_N - \inf f| \le \sup|f_N - f|\) and the triangle inequality give \[\left|d_{\rm St}(\widehat\mu_N, G \boxtimes \widehat T_N) - d_{\rm St}(\mu, G \boxtimes T)\right| \le d_{\rm St}(\widehat\mu_N, \mu) + d_{\rm St}(G \boxtimes \widehat T_N, G \boxtimes T) \le \varepsilon_N + \delta_N,\] uniformly over \(G \in \mathcal{H}\); taking suprema gives \(|\widehat\varepsilon_{{\rm model},N} - \varepsilon_{\rm model}| \le \varepsilon_N + \delta_N \to 0\) in probability. The argument uses only the triangle inequality, so it applies verbatim to the implemented discrepancy of Remark 2. \(\square\)

\emph{Mismatch.} If \(\mu \notin \mathcal{F}_T(\mathcal{H})\), the estimator targets the pseudo-true spectrum \(H^\dagger\). If \(H^\dagger\) is the unique minimizer and \(\kappa_{\rm inv}^{(\theta)}(H^\dagger, T;\mathcal{U}) > 0\), the argument of S-F applies with \(H^\dagger\) in place of \(H_0\); the forward residual at \(H^\dagger\) equals \(\varepsilon_{\rm model}\), which shifts the budget by an additive \(\varepsilon_{\rm model}\) in the forward metric only.

\section*{S-H. Proofs for Chapter 4 (legitimate rejuvenation)}\label{sm:S-H}\addcontentsline{toc}{section}{S-H. Proofs for Chapter 4 (legitimate rejuvenation)}

\emph{Lemma 5.} Metropolis--Hastings kernels satisfy detailed balance with respect to their actual target \(\widehat\pi_k^N\). With asymmetric proposals, for example when \(G_\beta\) enters the proposal, the full Hastings correction \(\alpha(x,x') = 1 \wedge \frac{\widehat\pi_k(x')q_k(x\mid x')}{\widehat\pi_k(x)q_k(x'\mid x)}\) restores reversibility. Hence \(\widehat\pi_k^N R_k = \widehat\pi_k^N\). The floor \(M_k \succeq \varepsilon I\) makes the proposal covariance nondegenerate in all directions; it does not by itself imply geometric ergodicity. \(\square\)

\emph{Lemma 6.} Under design choice I, \(\widehat\pi_k^N = \widetilde\pi_k^N\), so the move kernel leaves the conditional SMC target invariant. The extended-target representation of SMC samplers \cite{ref07} and of particle Markov chain Monte Carlo \cite{ref08} then applies: the selection-and-move particle system is exact on the extended target, and marginal consistency follows from Feynman--Kac semigroup theory \cite{ref04}, independently of the number of move steps. Under design choice II, Lemma 5 gives invariance only for the surrogate \(\widehat\pi_k\). \(\square\)

\emph{Theorem 7.} Work conditionally on \(\mathcal{F}_{k-1}^N\). Under the assumed drift--minorization condition for the resulting Metropolis--Hastings kernel, with \(\mathcal{F}_{k-1}^N\)-measurable constants \(C_k, \varrho_k\) (or, for preconditioned MALA, the cited regularity and tail assumptions \cite{ref30}), the conditional mixing bound \[d\Bigl(\mathcal{L}\bigl(X_{k,L_k}^{\rm alg}\,\big|\,\mathcal{F}_{k-1}^N\bigr),\ \widetilde\pi_k^N\Bigr) \le C_k\bigl(\mathcal{F}_{k-1}^N\bigr)\,\varrho_k\bigl(\mathcal{F}_{k-1}^N\bigr)^{L_k}\] holds in a metric compatible with the mixing theorem (total variation, a \(V\)-norm, or a weak metric dominated by it). The conditional triangle inequality then gives \[d\Bigl(\pi_k,\ \mathcal{L}\bigl(X_{k,L_k}^{\rm alg}\,\big|\,\mathcal{F}_{k-1}^N\bigr)\Bigr) \le d\bigl(\pi_k, \widetilde\pi_k^N\bigr) + C_k\varrho_k^{L_k}.\] For each fixed \(k\), standard Feynman--Kac stability assumptions imply \(d_{\rm BL}(\widetilde\pi_k^N, \pi_k) \to 0\) in probability; if the mixing bound is stated in a \(V\)-norm, the particle-filter term is matched in the same norm or dominated through it. Unconditional versions follow by taking expectations. Under design choice I the target-approximation term vanishes identically, and \(\widehat\Sigma^{\rm lift}\) enters only through the constants \(C_k\) and \(\varrho_k\). \(\square\)

\section*{S-I. Proofs for Chapter 4 (the full-cycle recursion)}\label{sm:S-I}\addcontentsline{toc}{section}{S-I. Proofs for Chapter 4 (the full-cycle recursion)}

\emph{Corollary 3.} This is the conjunction of Theorem 3 (the forward distortion) and Theorem 4 (the inversion budget). \(\square\)

\emph{Theorem 8.} The ideal distortion-free laws are defined by \(H_{k+1}^{\star,-} = (D_{a_k^2})_\# H_k^{\star,+} \boxplus \delta_{q_k}\) together with the ideal-correction identity \(H_k^{\star,+} := H_k^{\star,-}\); the algorithmic output is \(H_k^+ = \widehat H_k\), and the tracked error is \(\eta_k := W_1(H_k^+, H_k^{\star,+})\). For \(t \ge 0\), \(W_1((D_t)_\#\mu,(D_t)_\#\nu) = t\,W_1(\mu,\nu)\), and translation is an isometry, \(W_1(\mu \boxplus \delta_q, \nu \boxplus \delta_q) = W_1(\mu,\nu)\). With \(t = a_k^2\), propagation contributes the factor \(a_k^2 \le \bar a^2\). The distortion introduced by selection is removed by the correction step, whose error is bounded by \(\mathcal{B}_{k+1}^{\theta} = \max\{\theta, \mathcal{B}_{k+1}\}\) by Theorem 4 applied at step \(k+1\) with the uniform bound \(\kappa_{{\rm inv},k}^{(\theta)} \ge \kappa_0\). Iteration gives \(\eta_k \le \bar a^{2k}\eta_0 + \sum_{j<k}\bar a^{2(k-1-j)}\mathcal{B}_{j+1}^{\theta}\). The fixed-horizon statement needs only Theorem 4 at each step with \(\min_{k\le K}N_k \to \infty\); the long-horizon statement needs the time-uniform high-probability envelope \(\mathcal{B}_k \le b_k\), which a per-time \(o_p(1)\) statement does not imply. With the envelope, \(\limsup_k \eta_k \le \max\{\theta, \sup_k b_k\}/(1-\bar a^2)\), the fixed point of the recursion under the ceiling. For moment boundedness, corrected spectra lie in the common compact class, so \(m_2(H_k^+) \le b^2\), and the propagated laws obey \(m_2(H_{k+1}^-) = a_k^4 m_2(H_k^+) + 2a_k^2q_km_1(H_k^+) + q_k^2\), with first-moment recursion \(m_1(H_{k+1}^-) = a_k^2m_1(H_k^+) + q_k\); both are uniformly bounded under \(\bar a < 1\), \(q_k \le q_{\max}\), and the support bound. For positive invariance of \([a,b]\), the conditions \(a \le q_{\min}\) and \(b \ge q_{\max}/(1-\bar a^2)\) suffice, with the initial support contained in \([a,b]\). \(\square\)

\emph{Proposition 1.} In the setting of Theorem 8, let \(\mathcal{A} \subseteq \{1,\dots,K\}\) be the abstention set, and let the per-step budgets be \(b_j\) on corrected steps and \(\delta_{{\rm fb},j}\) on abstention steps, each capped below by \(\theta\) as in Section 3.4 of the main text. Substituting these budgets termwise in the error recursion gives \[\eta_K \le \bar a^{2K}\eta_0 + \sum_{j=1}^{K}\bar a^{2(K-j)}\Bigl[b_j\mathbf{1}_{\{j\notin\mathcal{A}\}} + \delta_{{\rm fb},j}\mathbf{1}_{\{j\in\mathcal{A}\}}\Bigr].\] Taking suprema, with \(\bar b^{\theta} = \max\{\theta, \bar b\}\) and \(\bar\delta^{\theta} = \max\{\theta, \bar\delta\}\), and separating the exchange term, \[\limsup_K \eta_K \le \frac{\bar b^{\theta}}{1-\bar a^2} + (\bar\delta^{\theta} - \bar b^{\theta})\,\limsup_K \Lambda_K, \qquad \Lambda_K \le \frac{1 - \bar a^{2A_K}}{1 - \bar a^2}\ \ (A_K = |\mathcal{A}|).\] If \(b_k \to 0\) and \(\Lambda_K \to 0\), the bound reduces to the resolution floor \(\theta/(1-\bar a^2)\). The relevant condition is a vanishing geometric load, not \(A_K = o(K)\): abstentions clustered at the end of the horizon are fully felt, since the geometric kernel has memory of order \((1-\bar a^2)^{-1}\). Finally, if \(H_k^{\star,+} \notin \mathcal{H}_{a,b}\), every output in \(\mathcal{H}_{a,b}\) lies at \(W_1\) distance at least \(\operatorname{dist}_{W_1}(H_k^{\star,+}, \mathcal{H}_{a,b}) > 0\) from the ideal law, which is positive because \(\mathcal{H}_{a,b}\) is closed in \(W_1\); when this distance exceeds \(\delta_{{\rm fb},k}\), abstention is the optimal feasible action. \(\square\)

\emph{Remark (accepted steps).} On accepted steps \(\widehat H_k \in \mathcal{H}_{a,b}\), hence \(\int\tau^2\widehat H_k(d\tau) \le b^2\) and \(\widehat H_k(\{0\}) = 0\) by the definition of the candidate class. This is a regularity property of accepted outputs, not a dynamical-stability statement.

\section*{S-J. Auxiliary computations}\label{sm:S-J}\addcontentsline{toc}{section}{S-J. Auxiliary computations}

\begin{enumerate}

\tightlist

\item

  Le Cam bound: \(d_{\rm TV}(\mathrm{Bin}(N, \lambda/N), \mathrm{Poi}(\lambda)) \le N(\lambda/N)^2 = \lambda^2/N\).

\item

  Poisson continuity in total variation: \(d_{\rm TV}(\mathrm{Poi}(\lambda), \mathrm{Poi}(\lambda')) \le |\lambda - \lambda'|\).

\item

  Multinomial factorial moments: \(\mathbb{E}[(n_{i_1})_{a_1}\cdots(n_{i_r})_{a_r}] = (N)_{a_1+\cdots+a_r}w_{i_1}^{a_1}\cdots w_{i_r}^{a_r}\) for distinct indices.

\item

  Linear-form variance: for \(\tilde n \sim \mathrm{Multinomial}(R; p)\), \(\operatorname{Var}(\sum a_i\tilde n_i) = R\left[\sum p_i a_i^2 - \left(\sum p_i a_i\right)^2\right] = \sum u_i a_i^2 - \left(\sum u_i a_i\right)^2/R\) with \(u_i = Rp_i\).

\item

  The maximum-atom rule for \(\boxtimes\) on \(\mathbb{R}_{\ge0}\) and the conventions of the companion fixed-point equation follow \cite{ref01}.

\item

  Poisson fourth moment: \(\mathbb{E}M^4 = \lambda^4 + 6\lambda^3 + 7\lambda^2 + \lambda\) for \(M \sim \mathrm{Poi}(\lambda)\), whence \(\operatorname{Var}(M^2) = 4\lambda^3 + 6\lambda^2 + \lambda\).

\end{enumerate}

\section*{S-K. Numerical conditioning on the two-point family}\label{sm:S-K}\addcontentsline{toc}{section}{S-K. Numerical conditioning on the two-point family}

\emph{Pipeline.} For two-point candidate spectra \(H = p_1\delta_{\sigma_1} + (1-p_1)\delta_{\sigma_2}\), \(\psi_H(z) = p_1\frac{\sigma_1 z}{1-\sigma_1 z} + (1-p_1)\frac{\sigma_2 z}{1-\sigma_2 z}\) is rational, so \(\chi_H = \psi_H^{-1}\) solves a quadratic equation; then \(S_H(w) = (1+w)\chi_H(w)/w\), \(S_{\mathrm{MP}(c)}(w) = 1/(1+cw)\), and \(S_\mu = S_H S_T\), with density recovery through the companion transform and Stieltjes inversion. For \(T = \mathcal{T}_c(\nu)\) the companion transform solves the fixed-point equation of \cite{ref01} with the validated solver.

\emph{Computed instance.} With \(d=900\), paired seeds, \(d_{\rm St}\) at bandwidth \(0.05\), and eight perturbations of \(H_0 = 0.7\delta_{0.5}+0.3\delta_{2.5}\): for the baseline \(T = \mathrm{MP}(c)\), that is \(\nu = \delta_1\), the estimated condition number declines mildly as \(c\) grows over the range studied, so pure sampling noise never collapses the conditioning. With resampling distortion, \(T = \mathcal{T}_c(\mathrm{MixPoi}(\rho_\alpha))\), the estimate is lowest at \((c,\alpha) = (0.75, 0.3)\), where the zero atom is largest, and stays of order one elsewhere. The individual values of this sweep are not registered in the deviation ledger; they are stated qualitatively here, and no sequence of them enters any theorem.

\emph{Map level and pipeline level.} The conditioning of the map itself degrades only mildly and tracks the zero-atom mass. The much stronger degradation measured in the experiments arises from the inversion pipeline (discretization, optimizer, and bandwidth) amplifying the ill-conditioning of the map, consistently with the bandwidth dependence of the activation boundary (S8). In short, the ill-conditioning is driven at the map level by the zero-atom mass and amplified at the pipeline level by discretization.

\emph{Status.} No explicit constant is claimed (S-E). The rigorous content is the positivity of Theorem 6. The computations of this section are numerical evidence on a single computable family: they evaluate closed-form transforms in floating point, without interval enclosures, and they serve to locate the degradation, at the map level or at the pipeline level, rather than to bound \(\kappa_{\rm inv}^{(\theta)}\). No quantity computed here enters any theorem.

\section*{S-L. Low-rank weight dependence leaves the spectral law unchanged}\label{sm:S-L}\addcontentsline{toc}{section}{S-L. Low-rank weight dependence leaves the spectral law unchanged}

\textbf{Proposition S2.} Let \(X_N = \Sigma_N^{1/2}Z_N\) under (G1)--(G3), and let the weights depend only on a \(q\)-dimensional projection, \(w_i = \omega(B^\top x_i)\) with \(B \in \mathbb{R}^{d\times q}\), \(q = q_N\) and \(q_N/d \to 0\). Then the limiting spectral distribution of the weighted or resampled second-moment matrix equals that of an exogenous-weight model with the same weight law: the dependence enters only through a channel of rank \(O(q)\), and the ESD deviation is \(O(q/d) \to 0\).

\emph{Proof.} Write \(x_i = m(u_i) + e_i\) with \(u_i = B^\top x_i \in \mathbb{R}^q\) and \(m(u_i) = \mathbb{E}[x_i \mid u_i]\), where \(e_i\) is independent of \(u_i\) with deterministic conditional covariance \(\Sigma_c\) (in the Gaussian case; in general this is the orthogonal decomposition onto the range of \(B\)). Condition on \(U = \{u_i\}\), which determines all weights. Then \[S_w \,\big|\, U = \frac1N\sum_i \lambda_i\, e_i e_i^\top \;+\; \frac1N\sum_i \lambda_i\bigl(m_i e_i^\top + e_i m_i^\top\bigr) \;+\; \frac1N\sum_i \lambda_i m_i m_i^\top, \qquad m_i := m(u_i).\] (i) Given \(U\), the weights are deterministic and the \(e_i\) are i.i.d. with covariance \(\Sigma_c\), so the first term is exactly an exogenous-weight model and Theorem 3 applies with the population law \(H_c\) of \(\Sigma_c\); since \(\Sigma - \Sigma_c\) has rank at most \(q\) by the Schur-complement form, \(|F^{H} - F^{H_c}| \le q/d\). (ii) The mean term has rank at most \(q\), each \(m_i\) lying in the range of \(B\). (iii) The cross term equals \(\frac1N(M\Lambda E^\top + E\Lambda M^\top)\) and has rank at most \(2q\). By the ESD rank inequality, \(\sup_x |F^{S_w} - F^{S_w^{\rm exog}}| = O(q/d) \to 0\). The conditional exogeneity required by (G4) is preserved for the resampling step, since counts given \((X,w)\) remain distributed according to the scheme with weights depending on \(U\) only. The moment controls needed are (W2) for the weight law, assumed on \(\omega\), and \(\mathbb{E}[\lambda\|m(u)\|^2] < \infty\). \(\square\)

\emph{Remark.} This explains why the mismatch certificate does not fire under low-rank endogenous weights (Section 5.3 of the main text): at the level of the limiting spectral distribution there is nothing to detect. The proposition partially relaxes (G4); general likelihood-driven weights remain out of scope (Section 6.1 of the main text).

\part*{Part II. Experimental sections}\addcontentsline{toc}{part}{Part II. Experimental sections}

\section*{S1. Count-law checks}\label{sm:S1}\addcontentsline{toc}{section}{S1. Count-law checks}

\begin{figure}[ht]
\centering
\includegraphics[width=0.9\linewidth]{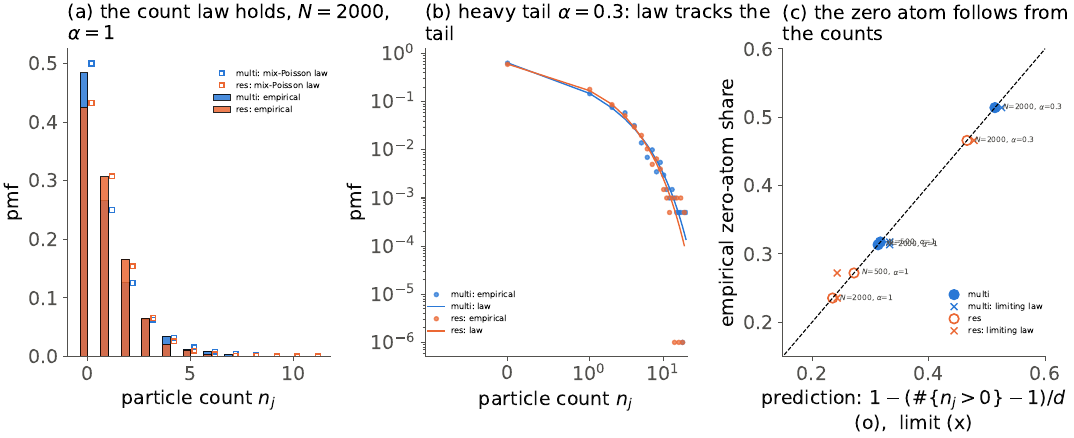}
\caption{Count-law checks. (a) Empirical probability mass function of the counts against the mixed Poisson law, $N=2000$, $\alpha=1$. (b) Heavy-tailed weights, $\alpha=0.3$, on logarithmic axes. (c) Empirical zero-atom share of the centered resampled matrix against the exact count-rank identity $1-(\#\{n_j>0\}-1)/d$ (circles) and the limiting value of Corollary 2 (crosses); all points lie on the diagonal.}
\label{fig:S1}
\end{figure}

The distribution of the count vector is the premise of the spectral-repair argument, and this section checks it. The figure reports the empirical probability mass function of the counts against the limiting laws (the mixed Poisson law for multinomial resampling, the floor-fraction law for residual resampling) for \(N \in \{500, 2000\}\), \(\alpha \in \{1, 0.3\}\), and both schemes. The mean of \(n_j\) equals \(1.00\) identically, as required by \(\sum_i n_i / N = 1\), and the \(L^1\) distances between the empirical and theoretical laws are \(0.028\) to \(0.060\), the magnitude of the sampling error expected at 2000 draws. Panel (c) places the source of the zero atom, the count-rank identity \(1 - (Np_s - 1)/d\), next to its cause, the count law: the zero atom is a consequence of bookkeeping, not a modeling assumption.

\section*{S2. Complete comparison for Corollary 2}\label{sm:S2}\addcontentsline{toc}{section}{S2. Complete comparison for Corollary 2}

\begin{figure}[ht]
\centering
\includegraphics[width=0.9\linewidth]{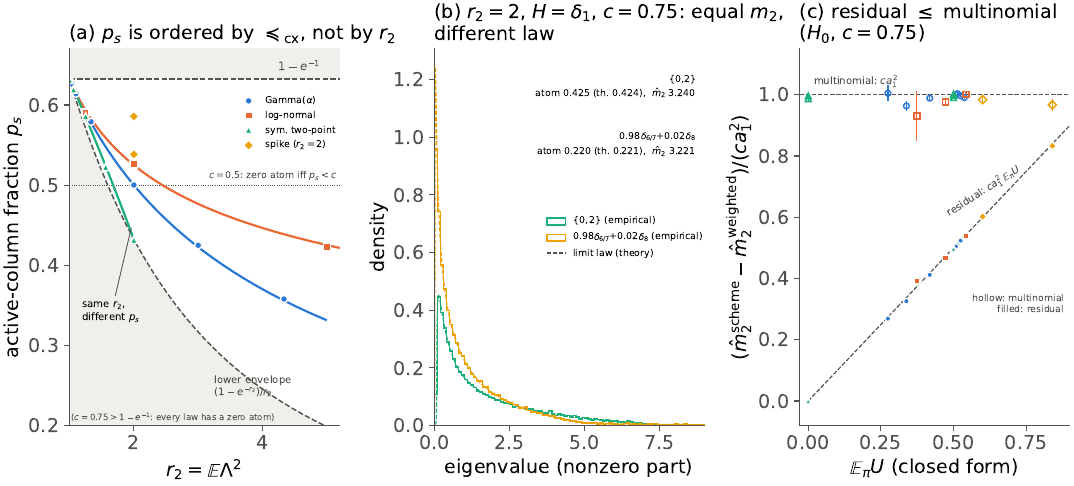}
\caption{Complete comparison for Corollary 2. (a) The active-column fraction $p_s$ is ordered by the convex order and not by $r_2$; equal $r_2$ can give different $p_s$. (b) Two laws with $r_2=2$ and equal second moment produce different limiting spectra and zero atoms. (c) The scheme-specific excess over the weighted ensemble: $ca_1^2$ for multinomial resampling and $ca_1^2\mathbb{E}_\pi U$ for residual resampling.}
\label{fig:S2}
\end{figure}

(1) The active-column fraction \(p_s\) is ordered by the convex order, not by the second moment. The three chains (Gamma with \(\alpha\) from \(100\) to \(0.3\), log-normal with variance from \(0.25\) to \(4\), and symmetric two-point with spread from \(0\) to \(1\)) decrease strictly as predicted, and all 17 laws fall within three standard errors plus \(0.002\) of the closed-form values; the upper bound \(1 - e^{-1}\) is attained only by the degenerate law. (2) The counterexample family at \(r_2 = 2\): five laws share the same second moment but differ in active-column fraction (from \(0.432\) to \(0.584\)), in zero-atom mass at \(c = 0.75\) (from \(0.220\) to \(0.425\)), and in the entire limiting law (pairwise closed-form \(W_1\) from \(0.030\) to \(0.185\)); at \(c = 0.5\) the difference becomes qualitative, with one law rank-deficient and two at full rank. This is the numerical evidence that \(r_2\) alone is not a strong enough partial order for \(p_s\). (3) The comparison of residual and multinomial resampling: the multinomial excess over the weighted ensemble is \(ca_1^2\) for all 17 laws, and the residual excess matches the closed form \(ca_1^2 \mathbb{E}_\pi[U]\) pointwise. With integer-valued weights, residual resampling equals the weighted ensemble exactly at every repetition, as it must.

\emph{Failed calibration test and its mechanism.} The pre-registered calibration test for the centered statistic, a cross-law regression of the centered second spectral moment on \(r_{2,N}\), fails: the measured slope deviates by about 8\% against a 2\% tolerance (slope \(0.834\) against the theoretical \(0.909\)). The post hoc diagnosis, recorded as such, is that the rank-one centering gap grows with the heaviest weight, and heavy-tailed weight laws (for example the log-normal law with variance \(4\), whose realized \(r_{2,N}\) ranges from \(3.5\) to \(26.5\)) make that gap non-negligible; the uncentered version of the same statistic is unbiased (mean error \(0.002\), standard deviation \(0.066\)). The linear law of Corollary 2(4) of the main text, which is exact conditionally on the counts, is therefore not refuted; what fails is the use of the centered statistic as a direct calibration under heavy-tailed weights. The evidence for Corollary 2(4) in the main text accordingly rests on items (2) and (3).

\section*{S3. Inversion sensitivity}\label{sm:S3}\addcontentsline{toc}{section}{S3. Inversion sensitivity}

\begin{figure}[ht]
\centering
\includegraphics[width=0.9\linewidth]{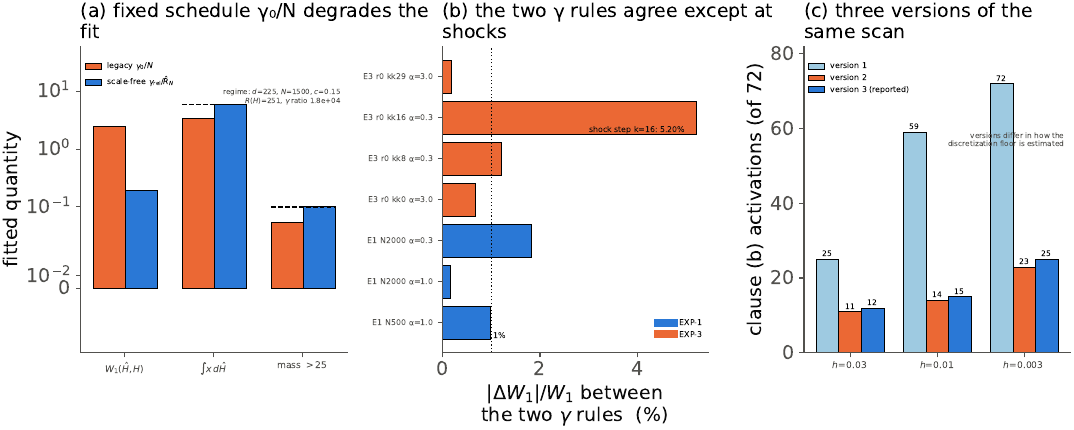}
\caption{Inversion sensitivity. (a) The fixed regularization schedule $\gamma_0/N$ against the scale-free schedule in the strong-anisotropy regime; dashed lines mark the true values. (b) Relative difference in $W_1$ between the two schedules, per experiment and step. (c) Activation counts of the comparative clause under three versions of the discretization-floor estimate, at three bandwidths; version 3 is the one reported.}
\label{fig:S3}
\end{figure}

Five items are reported. (1) The fixed-schedule failure case: a fixed schedule \(\gamma_N = \gamma_0/N\) does not adapt to the scale of \(H\), and in the strong-anisotropy regime the legacy rule degraded the fit (error \(W_1 = 2.51\) against \(0.19\) with the scale-free rule; recovered mean \(3.41\) against the true \(5.9\); spike mass \(0.053\) against the true \(0.1\)), with the legacy penalty and the data term of the same order (\(6.1 \times 10^{-5}\) against \(5.2 \times 10^{-5}\)): a penalty that ignores the problem scale is mis-set by an order of magnitude. (2) The comparison of the two rules: the maximum relative difference is \(5.20\%\) at the shock step of EXP-3, while EXP-1 at \(\alpha = 1\) shows only \(0.16\%\) to \(0.99\%\); this is the basis for adopting the scale-free rule throughout. (3) The fallback error against \(N\) (S4). (4) The comparison of the three plug-in variants (oracle, conditional, realized). (5) The solver validation table (Part III).

\section*{S4. Fallback error against N}\label{sm:S4}\addcontentsline{toc}{section}{S4. Fallback error against N}

\begin{figure}[ht]
\centering
\includegraphics[width=0.9\linewidth]{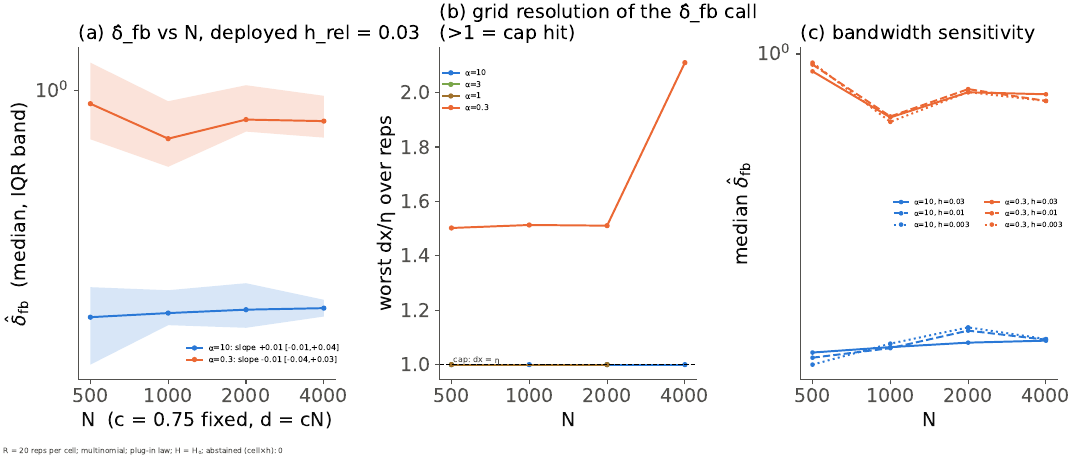}
\caption{Fallback error against $N$ at $c=0.75$. (a) Median $\widehat\delta_{\rm fb}$ with interquartile bands and bootstrap slope intervals, which contain zero for both weight profiles. (b) Worst grid-resolution ratio over repetitions; values above $1$ indicate that the resolution cap is reached. (c) Dependence on the bandwidth.}
\label{fig:S4}
\end{figure}

This experiment asks whether exceeding the grid ceiling affects the comparative clause of the supervisor. The design fixes \(c = 0.75\), which separates \(N\) from the aspect ratio since \(N = d/c\), with \(N \in \{500, 1000, 2000, 4000\}\), \(R = 20\), three Stieltjes bandwidths (\(0.03\), \(0.01\), \(0.003\)), and no abstentions. The fallback error does not decay with \(N\): all twelve slope confidence intervals contain zero (for the heavy-tailed law the slope is \(-0.0095\) with interval \([-0.0381, 0.0253]\)), and the medians are flat (\(0.974\), \(0.909\), \(0.944\), \(0.941\) across the four values of \(N\) for the heavy-tailed law). The registered conclusion is therefore that the bias does not decay with \(N\) and belongs to the intrinsic discretization of the count law, which supplies direct evidence for the absence of a closed-form finite-\(N\) correction. On the ceiling side, the heavy-tailed law exceeds the ceiling at every \(N\) tested (exceedance fractions \(10/20\), \(13/20\), \(17/20\), \(17/20\); worst ratio \(2.11\)), while the light-tailed laws never exceed it. The pre-registered non-monotonicity prediction was not reproduced: the worst ratio rises monotonically with \(N\), which indicates that the earlier observation was a per-repetition fluctuation rather than a stable effect of the integer structure.

\section*{S5. External baseline: Ledoit--Wolf nonlinear shrinkage}\label{sm:S5}\addcontentsline{toc}{section}{S5. External baseline: Ledoit--Wolf nonlinear shrinkage}

\begin{figure}[ht]
\centering
\includegraphics[width=0.9\linewidth]{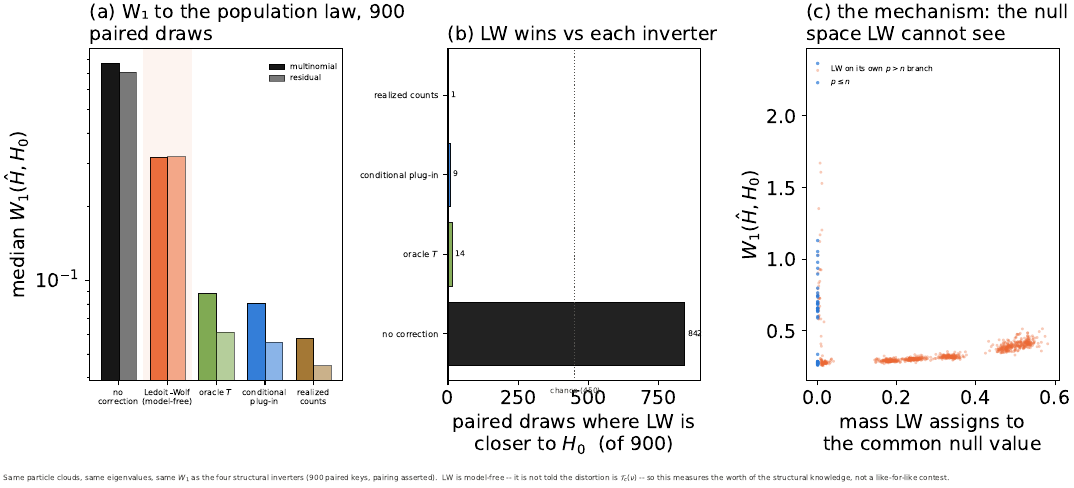}
\caption{Ledoit--Wolf nonlinear shrinkage as an external baseline. (a) Median $W_1$ error of each inverter, 900 paired draws. (b) Number of paired draws in which Ledoit--Wolf is closer to $H_0$ than each inverter. (c) $W_1$ error against the mass that Ledoit--Wolf assigns to the common null value.}
\label{fig:S5}
\end{figure}

Ledoit--Wolf nonlinear shrinkage \cite{ref15} is implemented as a fifth inverter on the same eigenvalues and the same paired clouds, with 900 paired assertions verified bitwise against the frozen EXP-1 rows (same particle clouds, same eigenvalues, same \(W_1\) convention). The baseline improves on the uncorrected spectrum (better in 842 of 900 cells) but falls short of every structure-aware inverter by a factor of \(4.3\) to \(6.2\): its median \(W_1\) is \(0.320\), against \(0.067\) for the conditional plug-in, \(0.074\) for the oracle, and \(0.052\) for the realized plug-in. The mechanism is visible in the data: 854 of 900 cells take the baseline's own \(p > n\) branch, which assigns uniform mass to the null space (median \(0.297\)), exactly where the count-rank identity restores the spectrum. The comparison is not like-for-like: the baseline is model-free and does not know that the distortion is the convolution with \(\mathcal{T}_c(\nu)\), whereas our inverters use that structure, and the gap quantifies the value of the structural knowledge.

\section*{S6. Phase diagrams and the supervisor audit table}\label{sm:S6}\addcontentsline{toc}{section}{S6. Phase diagrams and the supervisor audit table}

The phase diagram has two layers. The map level asks how identifiable the inverse problem is, through the median local condition number across the \((c, \alpha)\) grid; the pipeline level asks where the supervisor acts, through the activation fraction of the comparative clause. The deployment point \((c, \alpha) = (0.75, 3)\) is marked. Since the positive class of the abstention rule is empty on this deployment (no true positives in 660 steps), no ROC curve can be drawn, and the audit table instead reports the distribution of the supervisor's own decision ratio \(\widehat{\mathcal{B}}_N / \widehat\delta_{\rm fb}\): median \(0.146\), maximum \(0.824\), and no values above one. This is the measurable form of the statement that only a stronger degeneration would have triggered the clause.

\section*{S7. Metropolis--Hastings details}\label{sm:S7}\addcontentsline{toc}{section}{S7. Metropolis--Hastings details}

\begin{figure}[ht]
\centering
\includegraphics[width=0.9\linewidth]{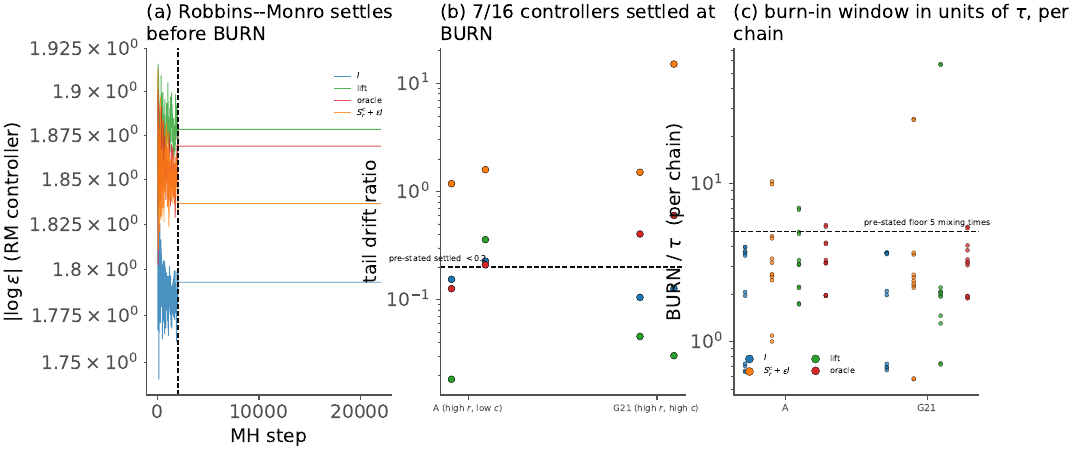}
\caption{Burn-in diagnostics for the Metropolis chains of Table 1. (a) The Robbins--Monro step-size controller settles before the end of burn-in (dashed). (b) Tail drift ratio of each controller at the end of burn-in, against the pre-stated threshold $0.2$. (c) Burn-in length in units of the integrated autocorrelation time, per chain, against the pre-stated floor of five.}
\label{fig:S7}
\end{figure}

This section reports the random-walk Metropolis details: effective sample sizes and integrated autocorrelation times, trace plots, the decay with chain length, the sweep over the \(\varepsilon\) floor, and the burn-in diagnostics. The burn-in budget is thin by the pre-stated reading rules (a drift ratio below \(0.2\) counts as settled, and a burn-in of at least five mixing times counts as adequate): the median burn-in window is between two and three mixing times, and 10 to 12 of the 12 chains fall below the five-mixing-time floor. The registered value \(1.943\) for the minimum burn-in window over the maximum mixing time is reproduced. With the adaptation frozen after burn-in, the production segment is a fixed-kernel Metropolis--Hastings chain, to which Theorem 7 applies directly; this is a property of the design, not empirical evidence of convergence.

\section*{S8. Activation zone of the comparative clause}\label{sm:S8}\addcontentsline{toc}{section}{S8. Activation zone of the comparative clause}

\begin{figure}[ht]
\centering
\includegraphics[width=0.9\linewidth]{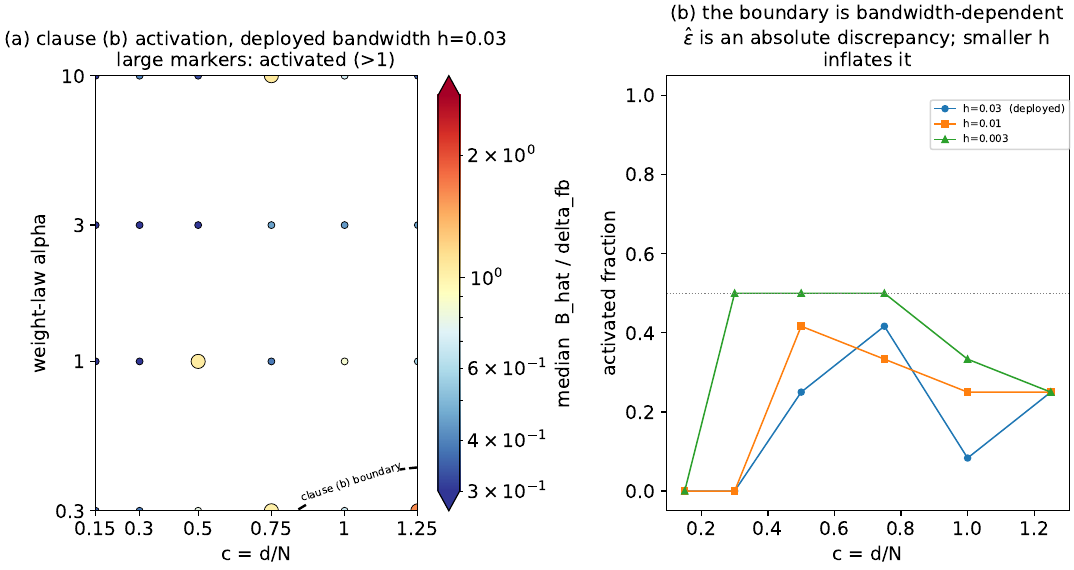}
\caption{Activation zone of the comparative clause. (a) Median decision ratio $\widehat{\mathcal{B}}_N/\widehat\delta_{\rm fb}$ on the $(c,\alpha)$ grid at the deployment bandwidth; large markers indicate activation. (b) Activated fraction against $c$ at three bandwidths.}
\label{fig:S8}
\end{figure}

This section maps the activation zone of the comparative clause across the \((c, \alpha)\) grid (six aspect ratios, four weight profiles, three repetitions) at three Stieltjes bandwidths. At the deployment bandwidth \(h = 0.03\,\xi_{\rm hi}\) the clause activates in 12 of 72 cells, concentrated at mid-to-high \(c\), with no activations at \(c = 0.15\) and \(c = 0.30\); the deployment point \((c, \alpha) = (0.75, 3)\) lies inside the boundary with a measured ratio of \(0.449 < 1\), so the clause is correctly inactive in EXP-3. The activation fractions are \(0.167\), \(0.208\), and \(0.347\) at the three bandwidths, reflecting the greater sensitivity of the discrepancy at smaller bandwidths (Section 2.1 of the main text). Two cells at \(c = 0.5\), \(\alpha = 1\) required the refinement ladder (the mass-capture guard was triggered, and refinement to finer smoothing restored it), and the corrected ratios are registered in the deviation ledger.

\section*{S9. Out-of-scope stress test (endogenous weights)}\label{sm:S9}\addcontentsline{toc}{section}{S9. Out-of-scope stress test (endogenous weights)}

This stress test lies outside the scope of Theorems 3 and 4 and is reported as such. Under likelihood-driven endogenous weights the mismatch certificate does not fire, both for a low-dimensional projection, \(q = 10\), where Proposition S2 shows that the limiting spectral law is unchanged and the inversion error indeed matches the permuted control, and for \(q = 750\), \(q/d = 0.5\), outside the scope of Proposition S2, where the median inversion error rises by a factor of \(1.22\). It fires on strongly correlated particles (autocorrelation \(0.9\), \(p = 4.4 \times 10^{-16}\)), and not at autocorrelation \(0.5\), where the error rises by a factor of \(1.35\). The test thus shows the boundary of the theory and the reach of the certificate: it detects the violations that inflate the inversion error severalfold, and misses milder ones.

\section*{S10. Passive monitor experiment}\label{sm:S10}\addcontentsline{toc}{section}{S10. Passive monitor experiment}

\begin{figure}[ht]
\centering
\includegraphics[width=0.9\linewidth]{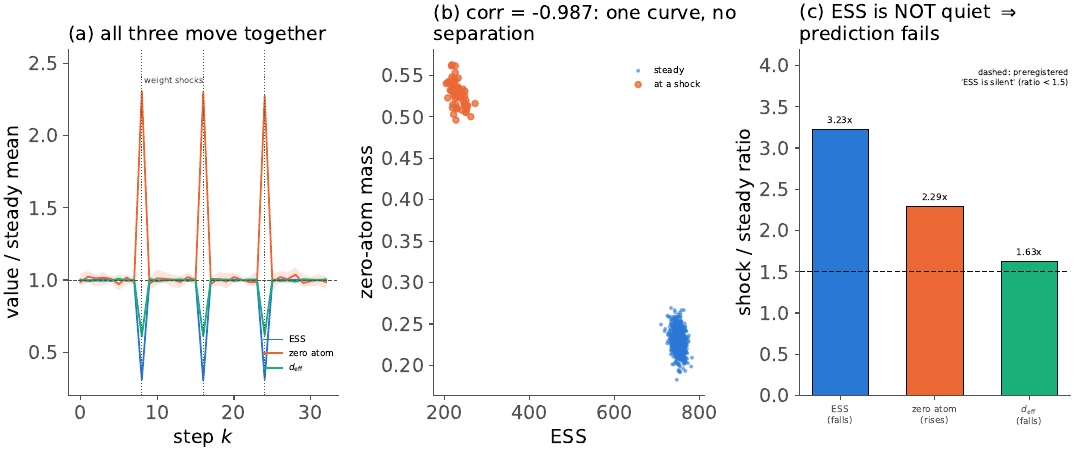}
\caption{Passive monitor experiment. (a) ESS, zero-atom mass, and effective dimension relative to their steady-state means; dotted lines mark the weight shocks. (b) Zero-atom mass against ESS; steady and shock steps form two separated clusters on one curve. (c) Shock-to-steady ratios against the pre-registered threshold $1.5$ for a silent ESS; the ESS falls by a factor of $3.23$, so the prediction fails.}
\label{fig:S10}
\end{figure}

The pre-registered prediction (entry M42) was that under a degeneration shock the spectral trajectories would raise an alarm while the effective sample size stayed silent. The measurement contradicts it: the effective sample size drops by a factor of \(3.23\) at the shock, in synchrony with the zero-atom mass (correlation \(-0.987\) at lag zero, with no detectable lead or lag), and the pairs with equal effective sample size but different spectral damage lie almost entirely on the steady-state noise floor (848 of 851), with only three pairs crossing a shock. By the pre-registered decision rule, the claim is weakened: spectral diagnostics complement the effective sample size rather than replace it, with a weak residual complementarity in the steady regime (the effective sample size explains only 6.8\% of the steady-state zero-atom fluctuation). The effective-sample-size blindness shown in \cite{ref01} is thus a static phenomenon, whereas dynamically the effective sample size moves together with the spectral damage. The figure shows the three trajectories with the shock markers.

\section*{S11. Theory--evidence correspondence}\label{sm:S11}\addcontentsline{toc}{section}{S11. Theory--evidence correspondence}

Table~\ref{tab:evidence} records the correspondence between the theoretical objects of the main text and the evidence bearing on them. ``Verified'' means the printed numbers test the stated object in the stated regime; ``trend'' means a mechanism is supported but not measured; ``estimate'' marks a number that is not a certified constant. A label of the form S$\cdot$ refers to a section of this supplement; all other numbers refer to the main text.

\begin{longtable}{@{}p{0.20\linewidth}p{0.15\linewidth}p{0.57\linewidth}@{}}
\caption{Theory--evidence correspondence.}\label{tab:evidence}\\
\toprule
Theoretical object & Evidence & Range supported \\
\midrule
\endfirsthead
\toprule
Theoretical object & Evidence & Range supported \\
\midrule
\endhead
\midrule
\multicolumn{3}{r}{Continued on next page}\\
\endfoot
\bottomrule
\endlastfoot

Theorem 1 (count laws) & Fig.~S1 & Count distribution, zero-atom origin and count moments (verified); not pointwise higher-order spectral detail. \\

Theorem 3, Corollaries 1--2 & Fig.~2(a,b) & Zero atom $0.33325$ against $0.33333$ and $0.24099$ against $0.24311$; second moments inside two standard errors in four of six groups, small negative bias in the two heavy-tailed groups (verified). \\

Theorem 4 (inversion and budget) & Fig.~2(c) & All three inverters lower the error by an order of magnitude with slopes bounded away from zero (verified); the implemented budget is an estimate, not a certified upper bound. \\

Theorem 5 (mismatch certificate) & Fig.~3(a) & Fires on the wrong mechanism and on strongly correlated particles, dormant for low-rank endogenous weights (verified as a boundary map); not an exogeneity test. \\

Theorem 6, Remark 3 (finite-resolution positivity) & Fig.~3(b) & Zero-atom mass tracks the condition number, Spearman $-0.931$ (trend); the modulus $\delta_\theta$ is neither measured nor certified. \\

Lemma 6, Theorem 7 (target preservation) & Table~1 & Condition number $42.4\to5.6$ with slow-direction ESS gains (verified); global null of no target displacement not rejected, \v{S}id\'ak-adjusted $p=0.059$. \\

Theorem 8 (full-cycle recursion) & Fig.~5(a) & Bounded trajectories recovering after shocks (verified); the fixed-$N$ plateau is the certified floor, not the error floor. \\

Proposition 1 (cost of abstention) & Fig.~5(b,c) & Geometric load $\Lambda_K=2.587$ against $0.459$ at $A_K=6$, $p=9.5\times10^{-7}$ (verified); the bound follows the load, not the count. \\

\end{longtable}

\part*{Part III. Protocols}\addcontentsline{toc}{part}{Part III. Protocols}

\section*{P1. Solver validation protocol}\label{sm:P1}\addcontentsline{toc}{section}{P1. Solver validation protocol}

The forward-map solver passes seven unit tests before any experiment: the correct branch (\(\operatorname{Im} m > 0\)); a fixed-point residual below \(3 \times 10^{-13}\) at interior and near-edge arguments; the normalization \(z m(z) \to -1\); recovery of the Marchenko--Pastur law within the reference tolerance; zero-atom masses consistent with the closed forms; insensitivity to the truncation of the count law; and cross-consistency between the covariance-side and companion-side implementations. Two further tests apply to the accelerated path: backend equivalence (deterministic quantities agree within the registered tolerance class, and chain-level quantities within statistical tolerance) and acceptance of the escalation ladder (each precision rung is checked pointwise, and typed events are exercised on deliberately constructed nonconvergent points). A mass-conservation check records the pre-normalization defect \(|\sum_j p_j - 1|\) in the run manifest, so that normalization cannot silently repair a defective density. Section 4.6 of the main text summarizes this protocol.

\section*{P2. Reproducibility protocol}\label{sm:P2}\addcontentsline{toc}{section}{P2. Reproducibility protocol}

The reviewer's path is a single command: clone the repository (release tag \texttt{v1.0-submission}) and run \texttt{python\ -m\ srlab\ run\ all}. Every printed number is regenerated from raw data through the four-level provenance chain (raw data, extraction function, intermediate quantity, printed value), and golden-regression checks compare any rebuilt result with the frozen baselines. The repository contains the full solver, the experiment modules, the deviation ledger with its 328 registered quantities, and the append-only pre-registration chain (entries M1 to M43). Data snapshots are archived with a DOI. Figures are regenerated from saved data only, and the PDF output is bitwise reproducible under a fixed \texttt{SOURCE\_DATE\_EPOCH}. The machine specification, library versions, and thread policy are recorded for each run in the manifests. Section 5.1 of the main text describes the pre-registration protocol that accompanies it.

\section*{P3. Spectrally corrected filtering step (complete specification)}\label{sm:P3}\addcontentsline{toc}{section}{P3. Spectrally corrected filtering step (complete specification)}

This section reproduces, as a numbered procedure, what Section~4.6 of the main text summarizes in prose. It is the single authoritative version; the main text carries no pseudocode.

\medskip

\textbf{Algorithm 1} (Spectrally corrected filtering step). Input: $X_{k|k-1}$, $w_{k-1}$, $y_k$, $g_k$, scheme flag.

\begin{enumerate}[label=(\arabic*),leftmargin=*]
\item \emph{Weight update.} $w_k^i \propto w_{k-1}^i g_k(y_k \mid x_k^i)$, normalized; record $\lambda_i = N w_k^i$ and $r_{2,N} := \frac1N \sum_i \lambda_i^2$.
\item \emph{Trigger.} If $r_{2,N} \le r_2^\star$, skip steps (3)--(7).
\item \emph{Selection.} Draw $n$ under the chosen scheme, subject to the integer-boundary protocol: integer parts are stored as integers, $R_N$ is formed by exact arithmetic, and $\lambda_i$ is never reconstructed from the floating-point product $N w_i$.
\item \emph{Spectral estimation.} Form $\widehat\mu_N$, with atoms reported separately from the density.
\item \emph{Distortion law.} Build $\widehat\nu_N$ and evaluate the companion transform of $\widehat T_N$ on the precomputed grid.
\item \emph{Regularized inversion.} Solve the Tikhonov problem with $\gamma_N = \gamma_{\rm rel}/\widehat R_N$; record $\widehat\varepsilon_{\rm model}$ and $\widehat\kappa_{\rm inv}$.
\item \emph{Lift.} Set $M_k = U_k \operatorname{diag}(\widetilde h_j) U_k^\top + \varepsilon I$, with $\widetilde h_j$ the ordered quantiles of $\widehat H_\gamma$.
\item \emph{Rejuvenation.} Run $L_k$ random-walk Metropolis steps targeting $\widetilde\pi_k^N$, with the damped transport used only as a proposal mechanism and the full Hastings correction for asymmetric proposals.
\item \emph{Supervisor.} Abstain if $\widehat\varepsilon_{\rm model} > \tau$ or $\widehat{\mathcal B}_N > \widehat\delta_{\rm fb}$, with $\eta_N^{\rm disc}$ subtracted from $\widehat\varepsilon$ before amplification; log every trigger and abstention by reason code.
\end{enumerate}

\end{document}